\documentclass[letterpaper]{article} % DO NOT CHANGE THIS
\usepackage{aaai24}  % DO NOT CHANGE THIS
\usepackage{times}  % DO NOT CHANGE THIS
\usepackage{helvet}  % DO NOT CHANGE THIS
\usepackage{courier}  % DO NOT CHANGE THIS
\usepackage[hyphens]{url}  % DO NOT CHANGE THIS
\usepackage{graphicx} % DO NOT CHANGE THIS
\usepackage{natbib}  % DO NOT CHANGE THIS AND DO NOT ADD ANY OPTIONS TO IT
\usepackage{caption} % DO NOT CHANGE THIS AND DO NOT ADD ANY OPTIONS TO IT
\usepackage{algorithm}
\usepackage{algorithmic}

\usepackage{multirow}
\usepackage{subfig}
\usepackage{float}
\usepackage{multicol}

\usepackage{amsmath}
\usepackage{amsthm}
\usepackage{booktabs}
\usepackage{algorithm}
\usepackage{algorithmic}
\usepackage[switch]{lineno}
\usepackage{multirow}
\usepackage{graphicx}
\usepackage{subfig}
\usepackage{xcolor}

\usepackage{amsfonts}       % blackboard math symbols
\usepackage{nicefrac}       % compact symbols for 1/2, etc.
\usepackage{microtype}      % microtypography
\usepackage{xcolor}         % colors
\usepackage{graphicx}
\usepackage{tcolorbox}
\usepackage{multirow}    
\usepackage{amsmath,amssymb,amsfonts}
\usepackage{subfig}
\usepackage{pifont}

\usepackage{overpic}
\usepackage{array} % for the 'm' column specifier
\usepackage{tabularx}
\usepackage{ltablex}
\newcolumntype{b}{X}
\newcolumntype{s}{>{\hsize=2.2\hsize}X}

\usepackage{newfloat}
\usepackage{listings}
\DeclareCaptionStyle{ruled}{labelfont=normalfont,labelsep=colon,strut=off} % DO NOT CHANGE THIS
\floatstyle{ruled}
\newfloat{listing}{tb}{lst}{}
\floatname{listing}{Listing}
\title{Multi-Modal Instruction-Tuning Small-Scale Language-and-Vision Assistant for Semiconductor Electron Micrograph Analysis}
\author{
    Sakhinana Sagar Srinivas\textsuperscript{\rm 1}\thanks{Designed, programmed the software, and drafted manuscript.},
    Geethan Sannidhi\textsuperscript{\rm 2}\thanks{Conducted experiments and analyzed visual results}, 
    Venkataramana Runkana\textsuperscript{\rm 1}\\
}
\affiliations{
    \textsuperscript{\rm 1} TCS Research,  %
    \textsuperscript{\rm 2}IIIT Pune \\
    \texttt{sagar.sakhinana@tcs.com}, \texttt{geethansannidhi20@cse.iiitp.ac.in}, \texttt{venkat.runkana@tcs.com}\\
}

\usepackage{bibentry}
\begin{document}

\maketitle

\begin{abstract}
\vspace{-2mm}
We present a novel framework for analyzing and interpreting electron microscopy images in semiconductor manufacturing using vision-language instruction tuning. The framework employs a unique teacher-student approach, leveraging pre-trained multimodal large language models such as GPT-4 to generate instruction-following data for zero-shot visual question answering (VQA) and classification tasks, customizing smaller multimodal models (SMMs) for microscopy image analysis, resulting in an instruction-tuned language-and-vision assistant. Our framework merges knowledge engineering with machine learning to integrate domain-specific expertise from larger to smaller multimodal models within this specialized field, greatly reducing the need for extensive human labeling. Our study presents a secure, cost-effective, and customizable approach for analyzing microscopy images, addressing the challenges of adopting proprietary models in semiconductor manufacturing.
\vspace{-5mm}
\end{abstract}

\vspace{-4mm}
\section{Introduction}
\vspace{-1mm}
The production of semiconductors is a complex, multi-step process involving specialized companies. Design firms, like Qualcomm, AMD, and NVIDIA, specialize in chip design, creating intricate integrated circuits for a variety of applications. They outsource semiconductor manufacturing to specialized foundries such as TSMC, Samsung, and GlobalFoundries, which handle the complex, multi-patterning fabrication process that includes photolithography, doping, etching, and chemical vapor deposition, all essential for imprinting the circuitry onto silicon wafers. After fabrication, the semiconductor chips undergo thorough quality checks before being enclosed in protective cases, connected to external pins, and finally assembled into devices like microprocessors and memory chips for diverse electronic systems. In the semiconductor industry, the push for miniaturization, especially using sub-7nm technology (transistors under 7 nanometers), leads to smaller, more powerful, and energy-efficient devices, boosting electronic product capabilities. Overcoming challenges in manufacturing precision and quantum tunneling is vital for the economical, large-scale production of highly advanced semiconductor devices using sub-7nm technology. This depends heavily on integrating advanced imaging for precise visualization, thorough testing for quality assurance, and innovative engineering. In the semiconductor industry, scanning electron microscopy (SEM) and transmission electron microscopy (TEM), are vital for precision. These tools provide high-resolution electron micrographs(microscopic images), essential for quality control, process optimization, and failure analysis. They enable precise characterization of semiconductor microstructures, ensuring products meet design specifications and facilitating improvements to reduce defects. This nanoscale characterization is key to technological advancement. Current technology faces challenges in complex material characterization within the semiconductor industry, particularly in the analysis of electron micrographs. Recent advances in AI, such as Large Multimodal Models (LMMs) like OpenAI's GPT-4 Turbo with Vision\cite{gpt4v}, combining language processing and visual understanding, are poised to significantly enhance semiconductor manufacturing. This technology is capable of analyzing high-resolution electron micrographs and identifying nanoscale structures and patterns, thereby aiding in quality control and improving manufacturing precision and efficiency. Proprietary LMMs like GPT-4 offer significant advantages but face adoption challenges, particularly due to concerns about sensitive data exposure that could risk revealing the designs and processes of semiconductor firms. In contrast, open-source, small-scale multimodal models (SMMs), such as LLaVA\cite{liu2023visual} and MiniGPT-4\cite{zhu2023minigpt}, provide a cost-effective, fine-tuning approach for domain-specific customization in nanomaterial image analysis. While these models are more interpretable, they may not match the reasoning and generalization capabilities of their proprietary counterparts, potentially leading to less coherent outputs. In addition, acquiring the high-quality data necessary for training SMMs in nanomaterial image analysis is difficult due to limited and expensive datasets. The need for expert knowledge and specialized tools for annotation adds to the cost and time. Moreover, the diversity in image characteristics from various material imaging techniques makes the development of a generalizable, one-model-fits-all approach very challenging across different electron micrograph datasets. In our study, we present a novel approach that utilizes GPT-4 Turbo with Vision, an advanced multimodal large language model, as a robust ``teacher" for generating instruction-following data, specifically question-answer pairs related to nanomaterial image analysis. Building on this innovative dataset, we present the Multimodal Vision Assistant for Electron Micrograph Analysis (\texttt{MVaEMa}), an end-to-end trained smaller multimodal model (SMM), designed to be more efficient while still being powerful. We instruct-tune \texttt{MVaEMa} using the aforementioned machine-generated dataset—a comprehensive collection of vision-language corpora for domain-specific customization. Each labeled pair includes a query image, a related text instruction, and the most accurate response or description. We utilize vision-language instruction tuning to enhance the zero-shot capabilities of the proposed framework, \texttt{MVaEMa}, for tasks like visual question answering(VQA) on nanomaterial image analysis, while adhering to the auto-regressive training approach, thus eliminating the need for high-quality, human-annotated image-text pairs for domain-specific adaptation. Training smaller models through vision-language instruction tuning using larger multimodal models is a promising approach, leveraging the knowledge and capabilities of the larger models. This method involves transferring knowledge from the larger model (the teacher) to the smaller model (the student) to enhance performance, enabling better understanding of visual concepts and accurate text generation based on visual content. This method improves grounded language generation and visual reasoning through the distillation of knowledge from teacher models, which is accomplished by aligning the student model's predictions with those of the teacher model. 

\vspace{-3mm}
\begin{figure}[ht!]
\centering
\resizebox{1.1\linewidth}{!}{ 
\hspace*{-5mm}\includegraphics[keepaspectratio,height=4.5cm,trim=0.0cm 0.0cm 0cm 0.5cm,clip]{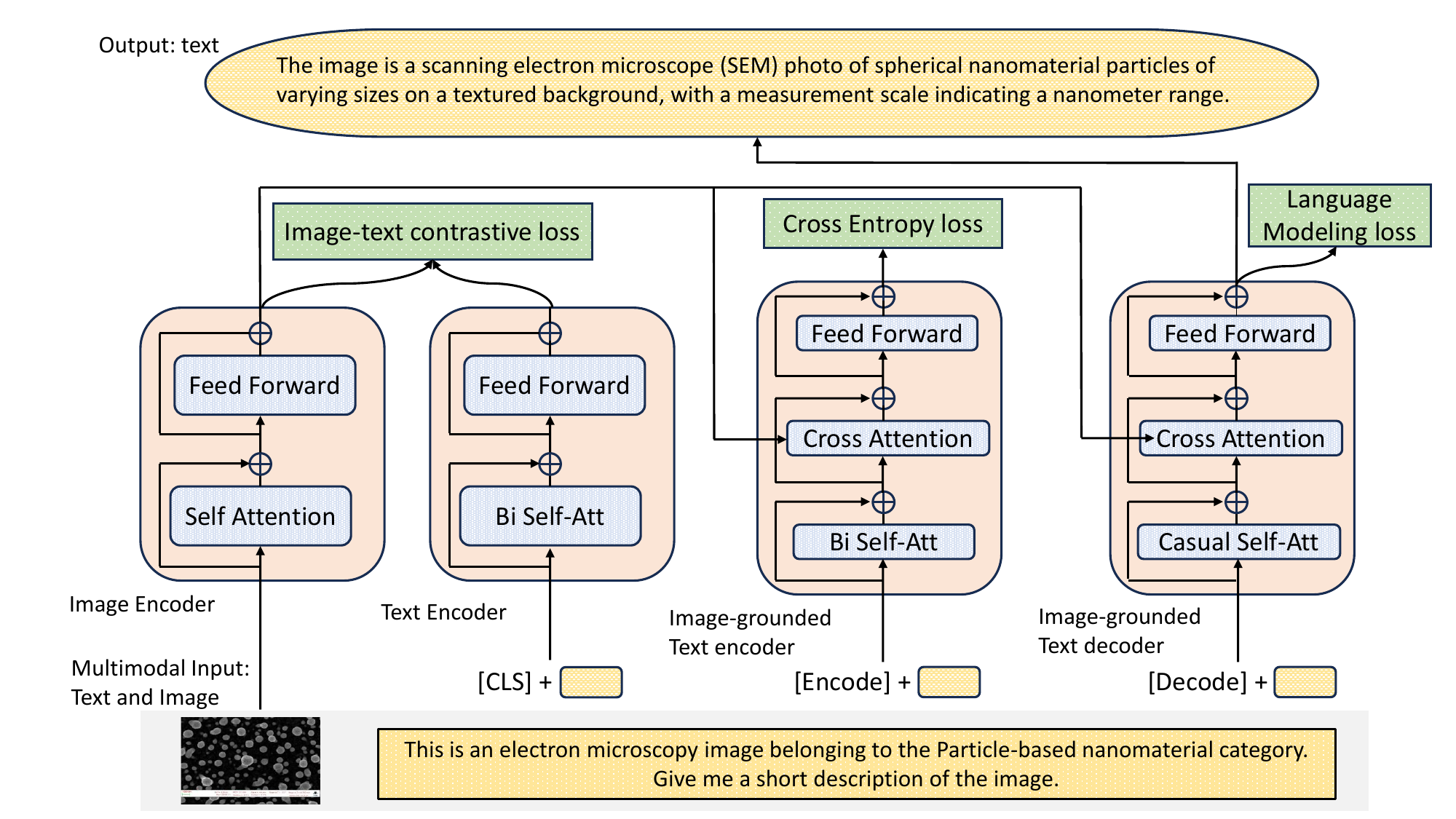} % left, bottom, right, top
}
\vspace{-6mm}
\caption{The architecture and objectives of \texttt{MVaEMa}, our proposed multimodal deep learning framework for VQA task in nanomaterial image analysis, are presented. The schematic illustrates a small-scale multimodal architecture that integrates text and image data, which is trained using vision-language instruction tuning, utilizing instruction-following data generated by the instruction-tuned GPT-4 Turbo with Vision. The architecture consists of an image encoder, a text encoder, and an image-grounded text-encoder and text-decoder, each containing self-attention and feed-forward layers. The framework is optimized using a combination of image-text contrastive, binary cross-entropy, and language modeling loss functions, aiming to align the multimodal representations to generate text output that answers questions about the image, showcasing the framework's ability to interpret and articulate complex intermodal relationships.
}
\label{fig:figure1}
\vspace{-4mm}
\end{figure}

\vspace{-1mm}
Furthermore, enterprises can fine-tune the proposed pretrained model, \texttt{MVaEMa} on their proprietary data within their infrastructure, thus ensuring privacy, reducing costs, increasing customization, and enhancing security. Overall, it presents a viable solution potentially democratizing access to their capabilities and accelerating their adoption for various multimodal tasks, aligning with the increasing need for personalized, private AI solutions. We present the architecture of the proposed framework, \texttt{MVaEMa}, in Figure \ref{fig:figure1} for the zero-shot visual question-answering task. The proposed framework is a small-scale, autoregressive, unified vision-language model that employs an encoder-decoder architecture to process and integrate both text and image modalities. The multimodal input consists of the query microscopic image and the corresponding natural language question (task instruction), with the goal of providing an accurate answer based on the image content. The multimodal model comprises the following components: (a) The instruction-aware \textbf{image encoder} uses a self-attention mechanism with a larger global receptive field to analyze visual inputs, capturing salient information, long-range dependencies and the overall scene composition. This allows the multimodal model to understand the global context of an image in a holistic and flexible manner, highlighting important regions and their contextual relationships while computing expressive image embeddings. (b) \textbf{The text encoder} is crucial for understanding and interpreting the query text, ensuring that it can be effectively combined with visual information for cross-modal analysis to provide accurate and relevant answers. The text encoder employs a bidirectional self-attention mechanism to encode linguistic inputs, preserving semantic and learning contextual relationships. We use a $\textless \textit{cls}\textgreater$ token to represent the entire sequence, providing a rich, contextualized representation of the query text essential for integrating with visual information to generate precise descriptions. The $\textless \textit{cls}\textgreater$  token embedding helps the multimodal model focus on relevant parts of the image and guides the answer generation process based on the question's context. The unimodal encoders (i.e., both text and image encoders) compute respective monolithic embeddings, which are jointly trained with a image-text contrastive loss to align the vision and language embeddings. (c) \textbf{The image-grounded text encoder} employs an additional cross-attention mechanism to align specific textual information with relevant visual features, computing contextually relevant multimodal representations.  We utilize binary cross-entropy loss in image-text matching to assess a multimodal model's ability to correctly match images with text, aiming to minimize the discrepancy between positive and negative image-text pairs. This process results in precise, context-aware textual descriptions that accurately reflect the visual information. (d) \textbf{The image-grounded text decoder} utilizes the rich, multimodal representations to generate a syntactically and semantically coherent, contextually relevant textual description corresponding to the visual input. The decoder replaces the bi-directional self-attention layers with causal self-attention and employs the same cross-attention layers and feed-forward networks as the image-grounded text encoder for text generation. It is trained with a language modeling loss to produce an output description that accurately reflects the image's content and context, thereby bridging the gap between visual perception and language generation by grounding the output in the image's visual content. The multimodal framework is optimized using a combination of image-text contrastive, cross-entropy, and language modeling loss functions, ensuring alignment between modalities and linguistic accuracy. This sophisticated approach enables the framework to answer questions about images with a high degree of precision and relevance. We train our multimodal framework using a specific type of instruction-following data: VQA task-based image-instruction-answer pairs. Based on this machine-generated data, we design a multimodal prompt to customize the \texttt{MVaEMa} framework, where the objective is to analyze the query image and provide an accurate answer based on the visual content and the specific question. As depicted in Figure \ref{fig:figure1}, we adopt a symbolic approach with prompting mechanism, wherein the prompt( i.e., caption + natural language instruction), the caption explicitly mentions the microscopy image belongs to the predefined nanomaterial category (ground-truth). This description serves as a symbolic representation that the language encoder recognizes, and it decodes this sequence to understand the visual information from an SEM image of nanomaterials. Consequently, it integrates the image information with linguistic context within the multimodal model's processing framework. 

\vspace{-4mm}
\begin{figure}[htbp]
     \centering
     \subfloat[High intra-class dissimilarity(variance) in electron micrographs of a nanomaterial (\textit{micro-electromechanical systems(MEMS)} device).] 
     {\includegraphics[width=0.12\textwidth]{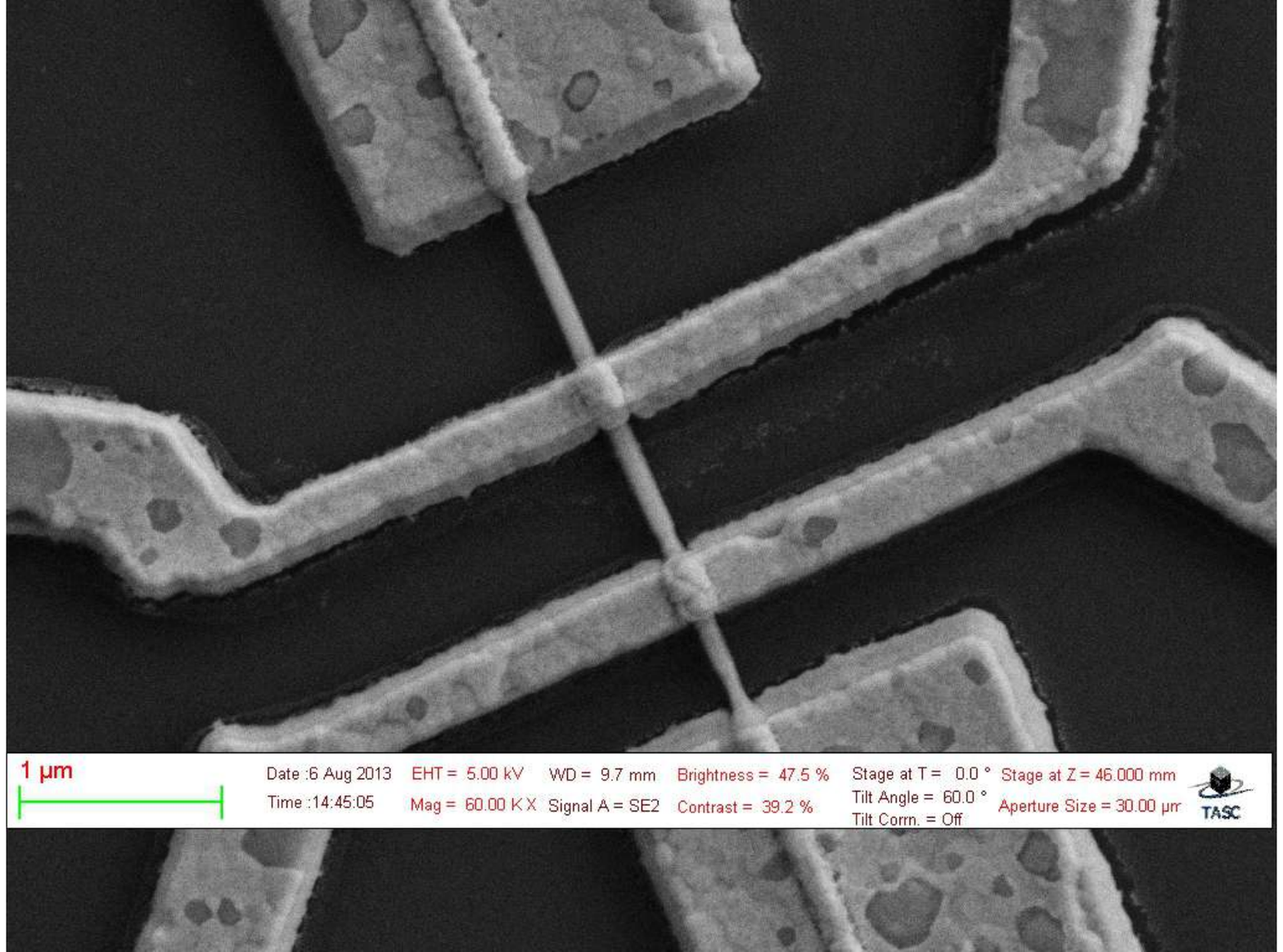}
     \includegraphics[width=0.12\textwidth]{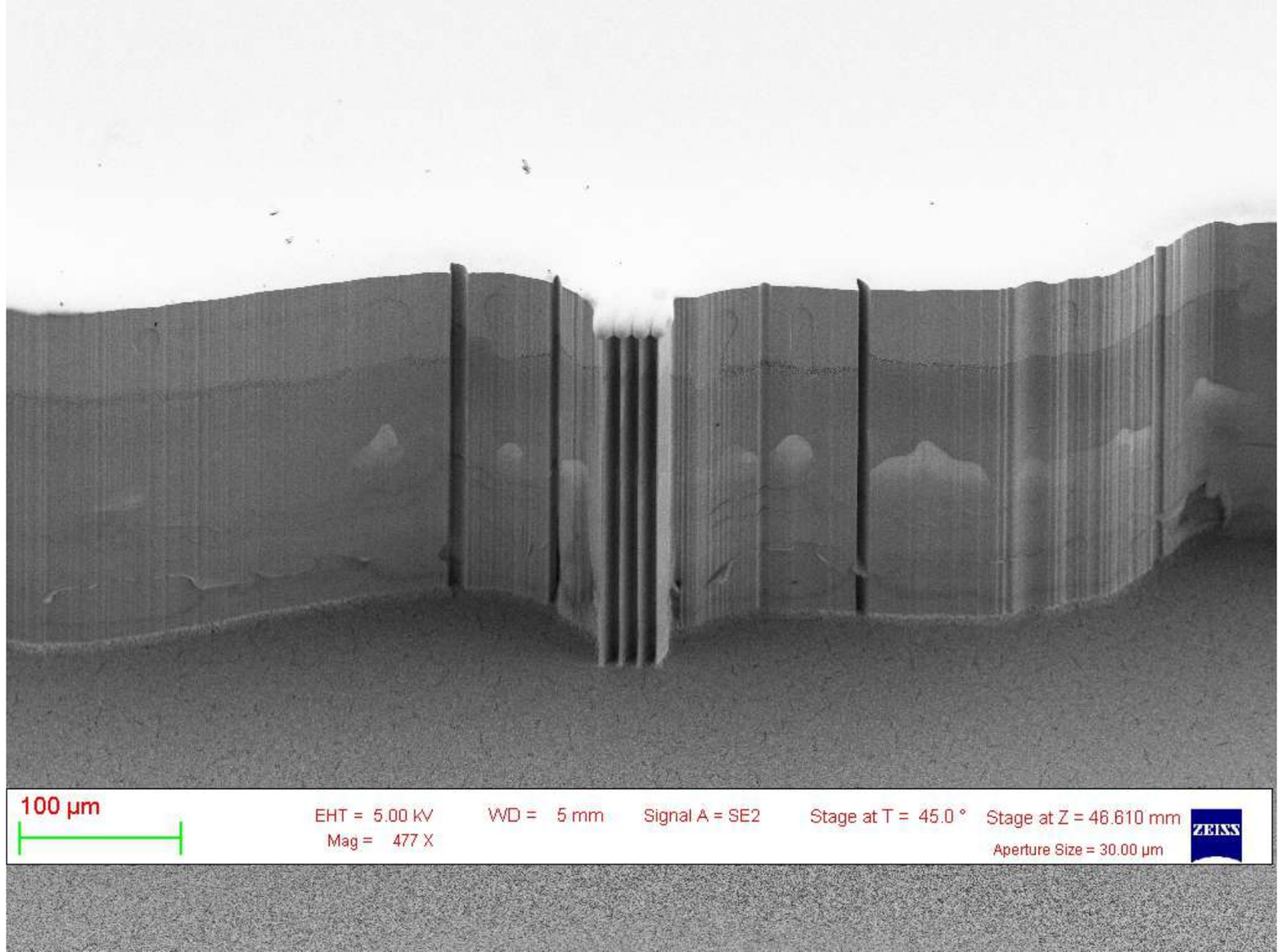}
     \includegraphics[width=0.12\textwidth]{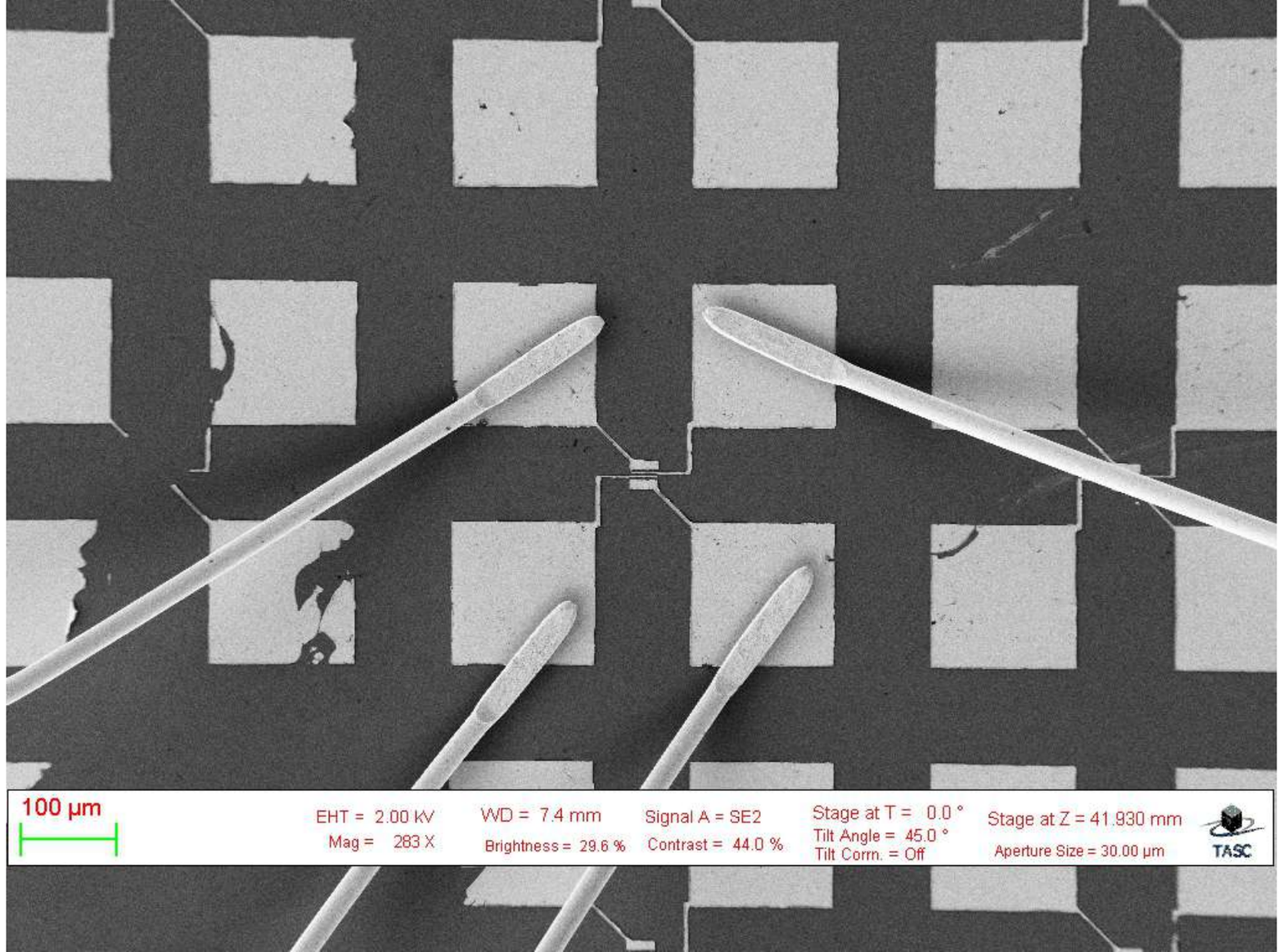}
     \includegraphics[width=0.12\textwidth]{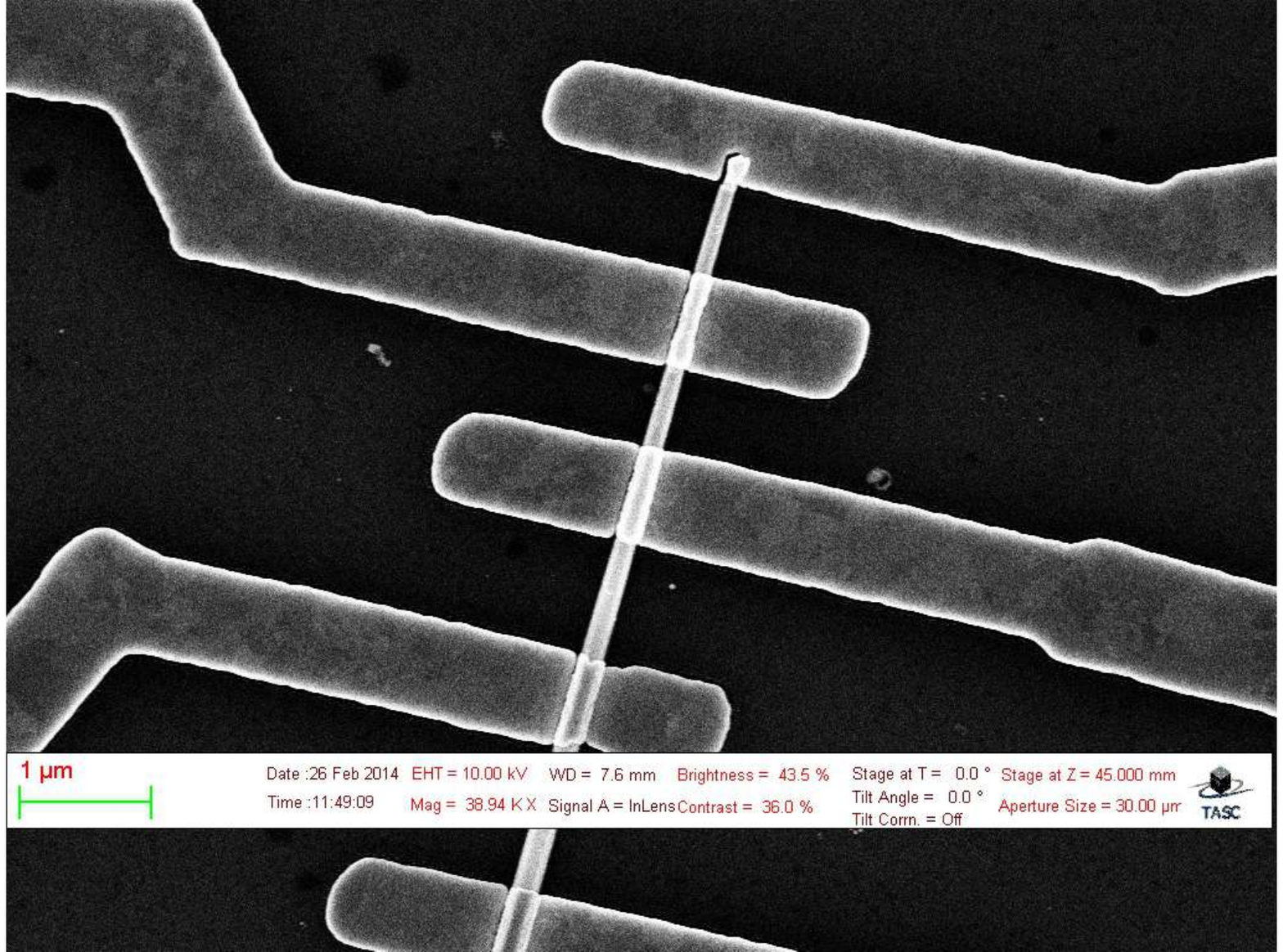}
     }
     \vspace{-0.5mm}
     \qquad
     \subfloat[High inter-class similarity: Electron micrographs of different nanomaterials (\textit{porous, particles, powders, films}) show noteworthy similarity.]{\includegraphics[width=0.12\textwidth]{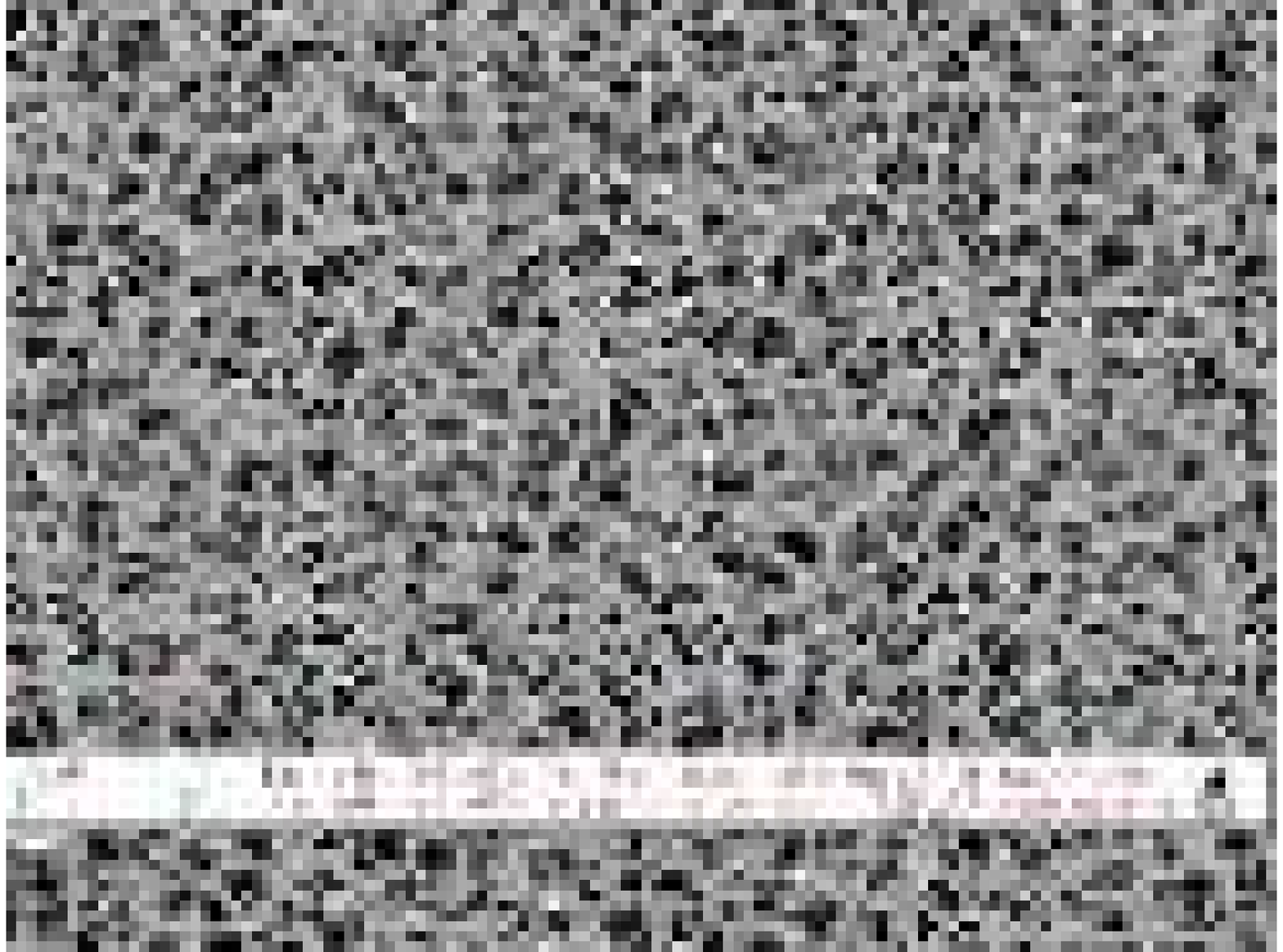}
     \includegraphics[width=0.12\textwidth]{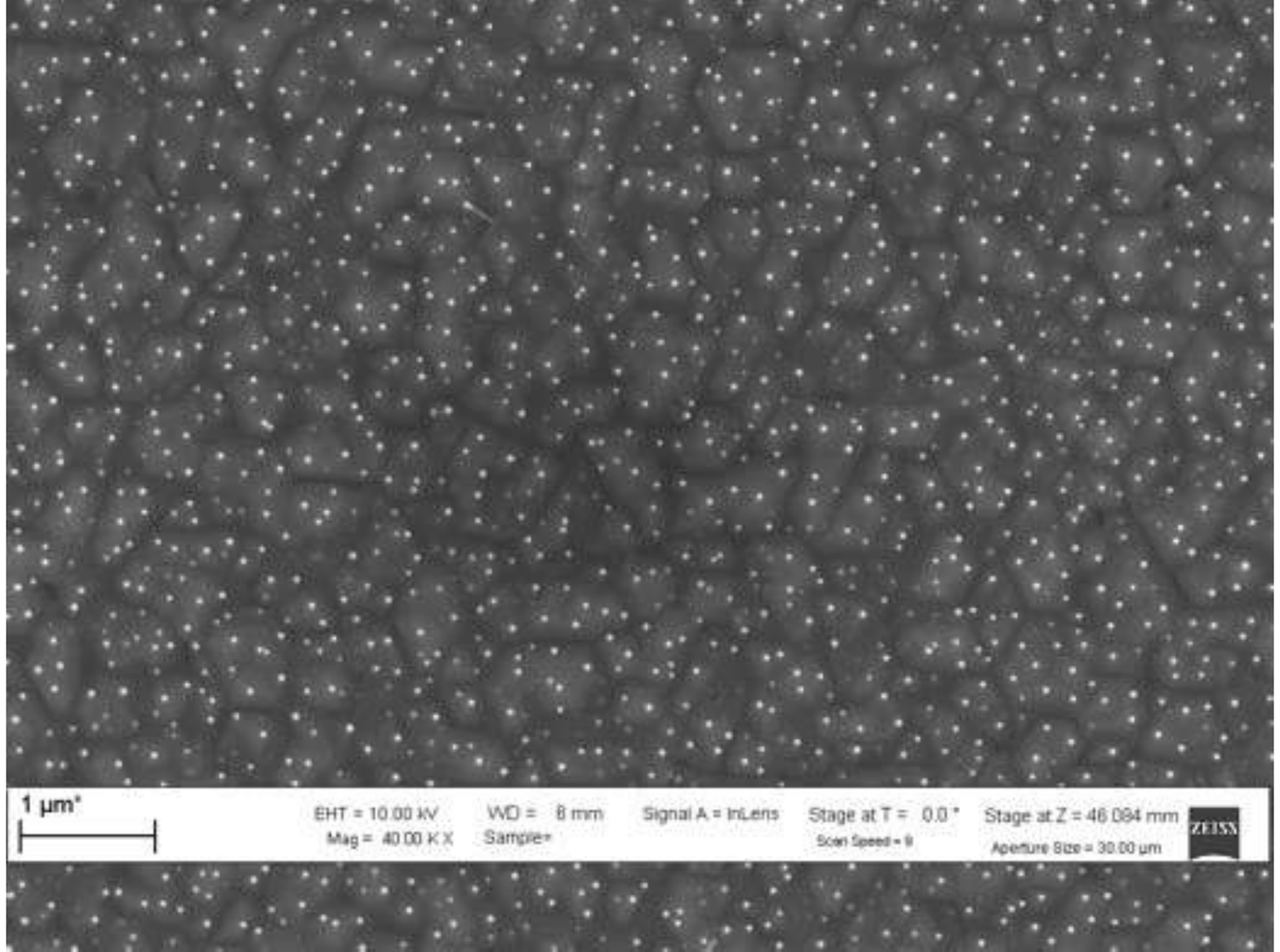}
     \includegraphics[width=0.12\textwidth]{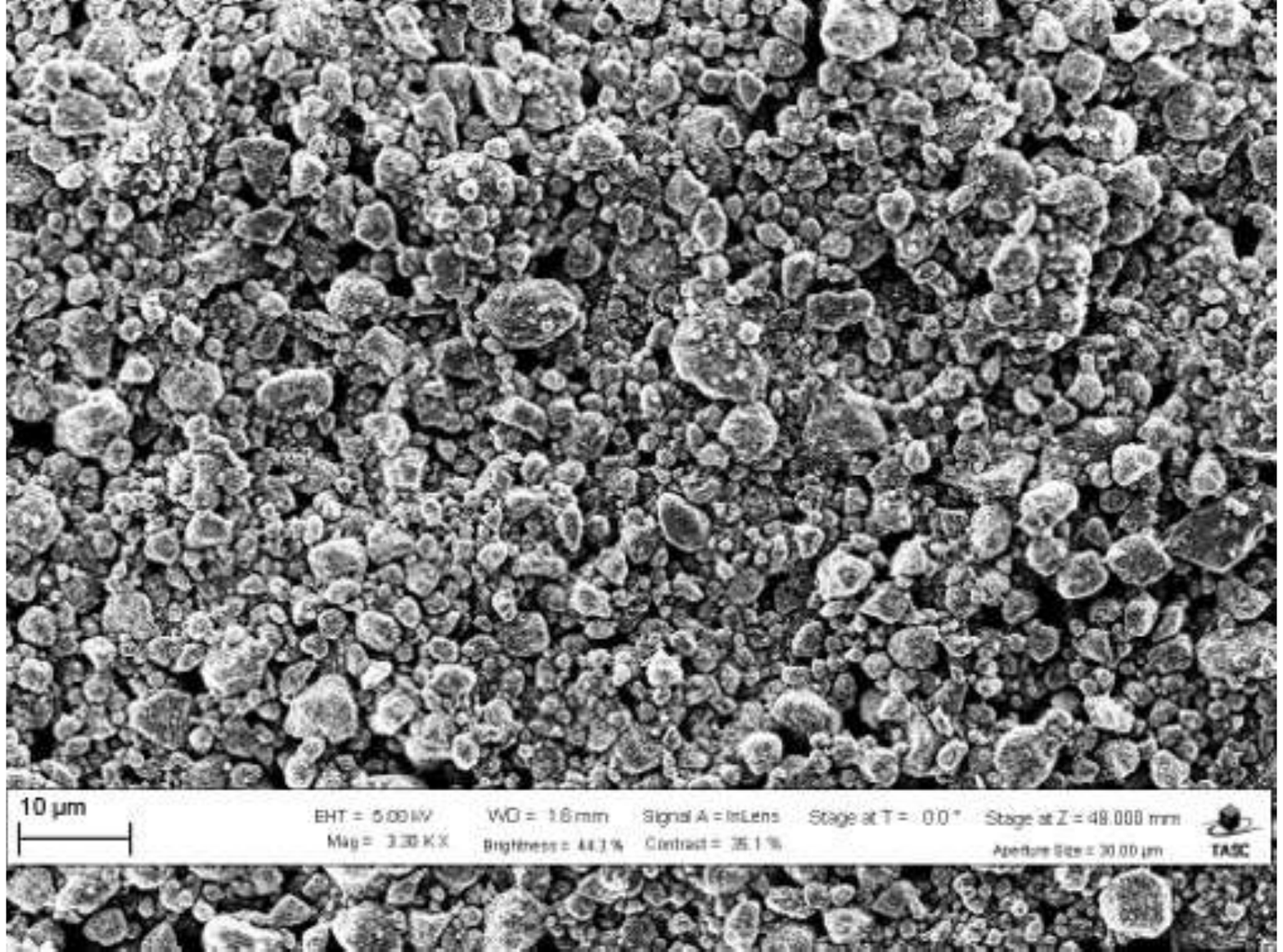}
     \includegraphics[width=0.12\textwidth]{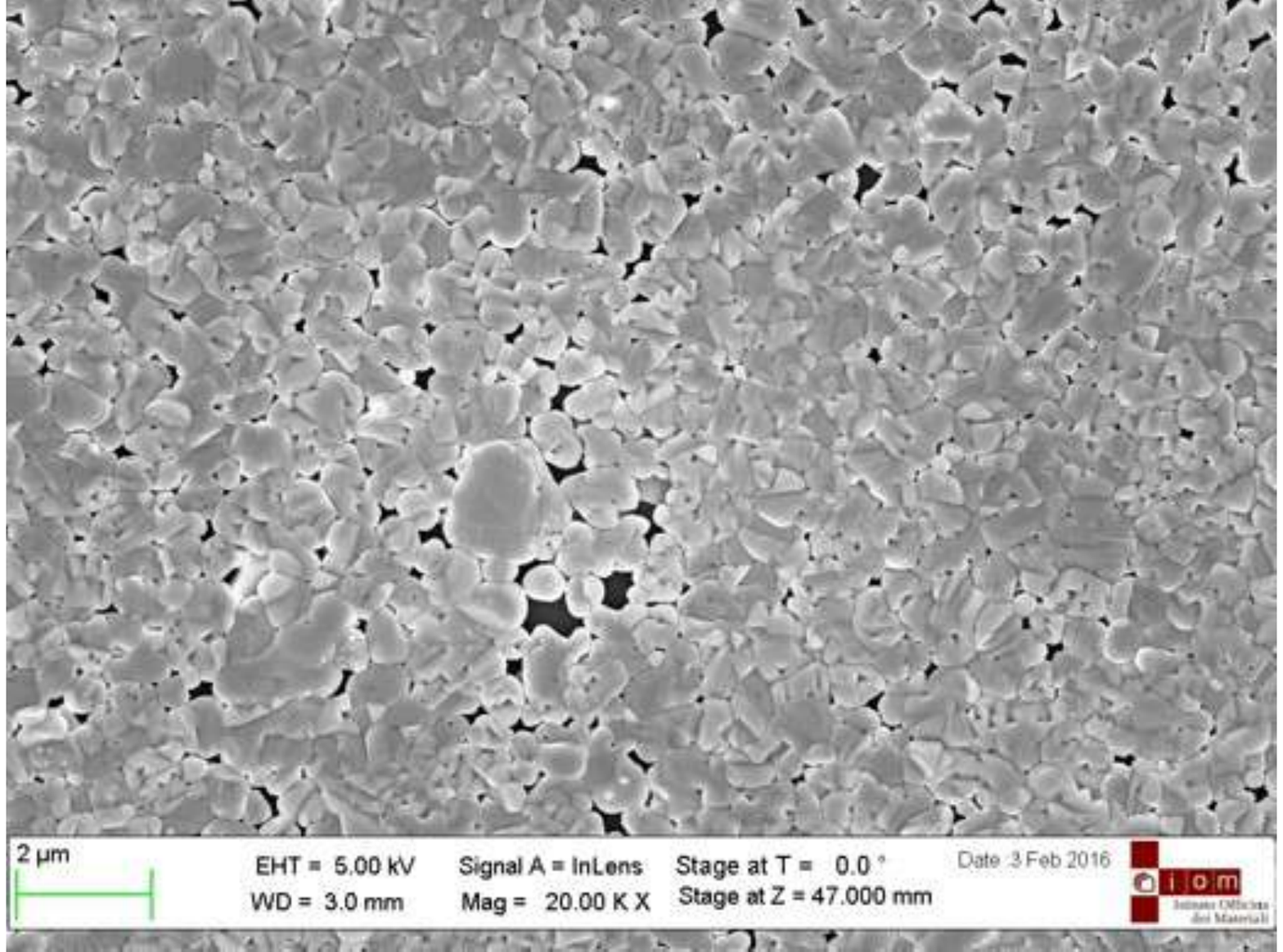}
     }
     \vspace{-0.5mm}
     \qquad
     \subfloat[Multi-spatial scales of patterns: Nanoparticle electron micrographs exhibit multi-scale spatial heterogeneity.]
     {\includegraphics[width=0.12\textwidth]{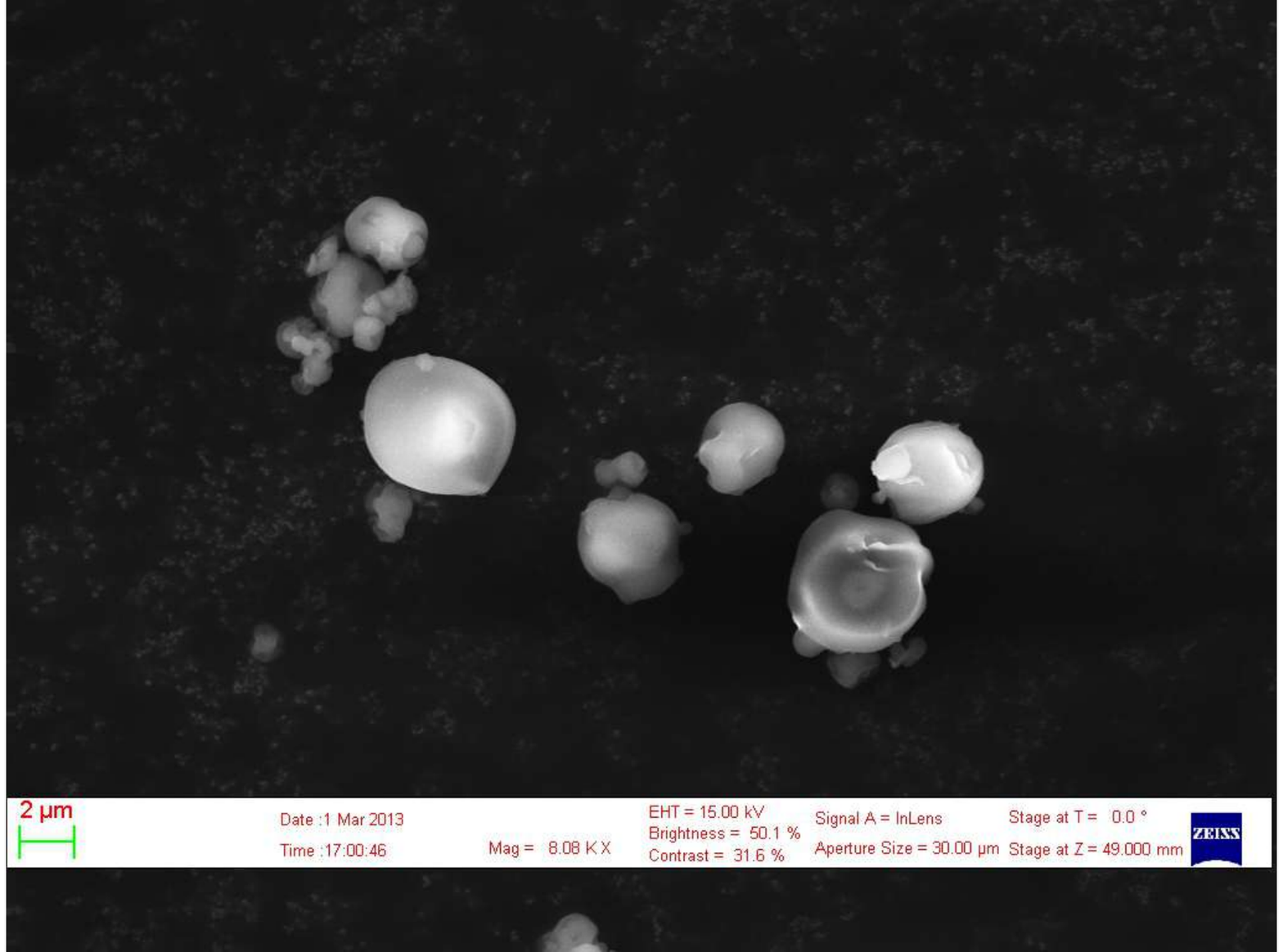}
     \includegraphics[width=0.12\textwidth]{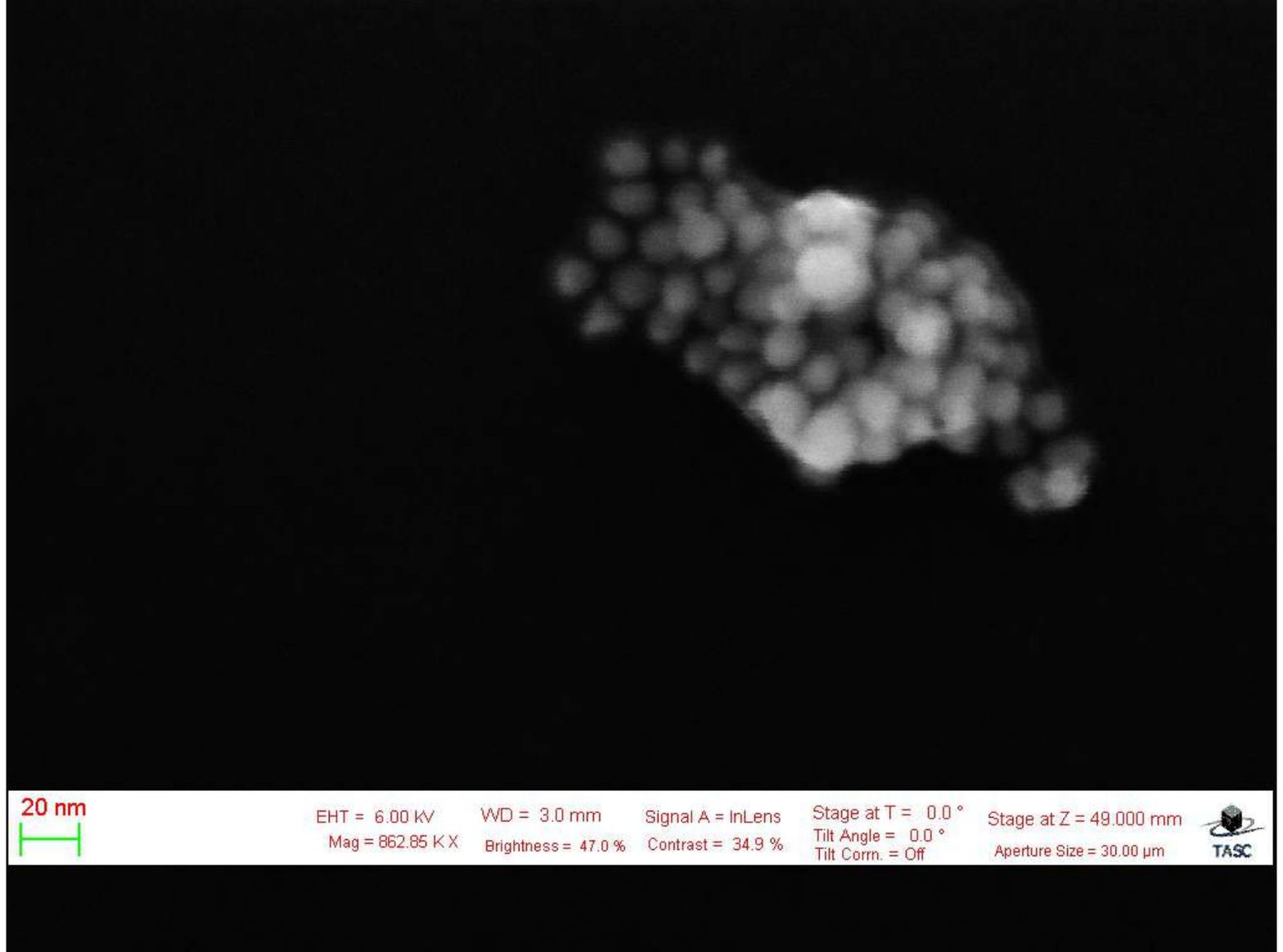}
     \includegraphics[width=0.12\textwidth]{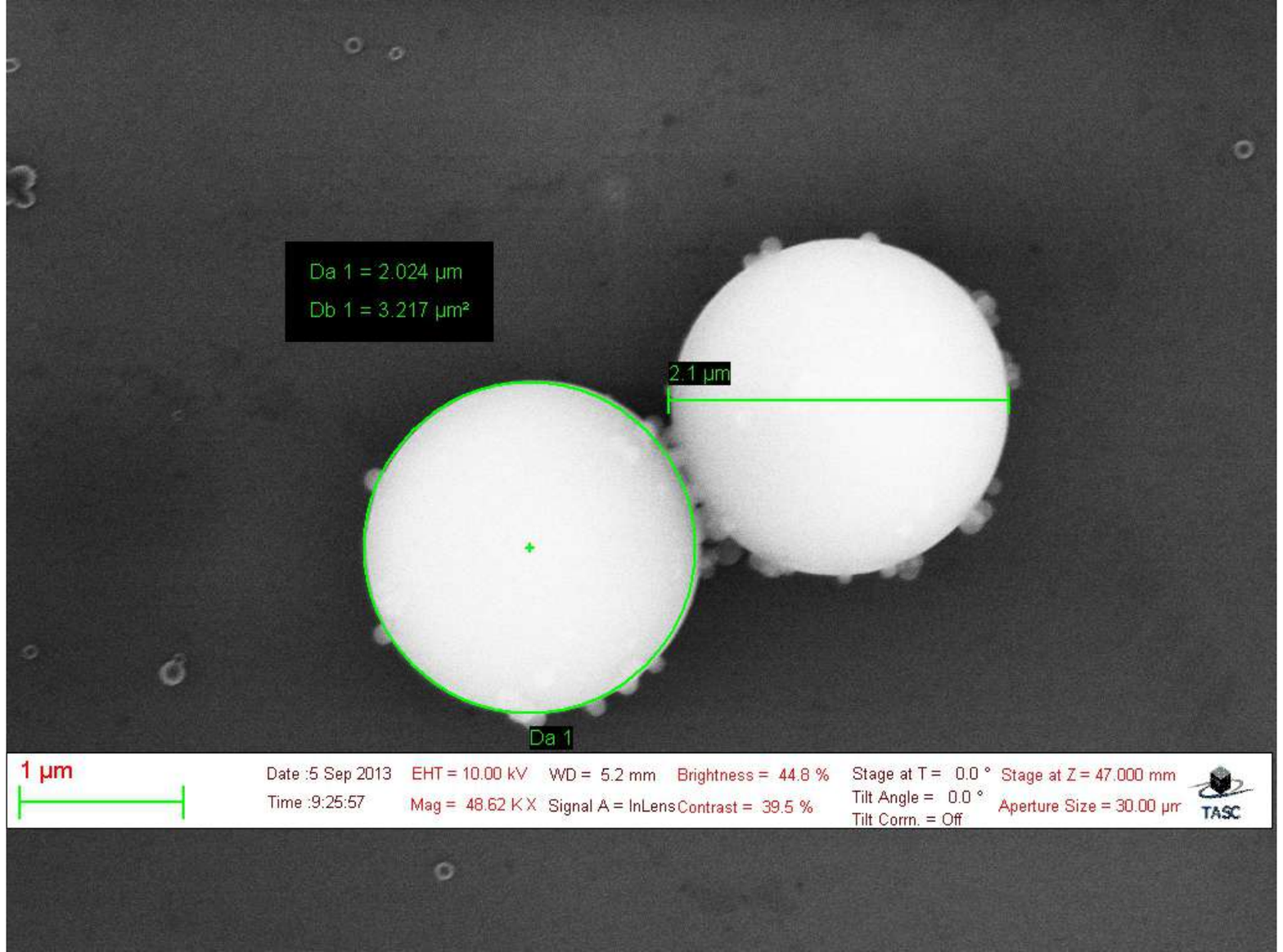}
     \includegraphics[width=0.12\textwidth]{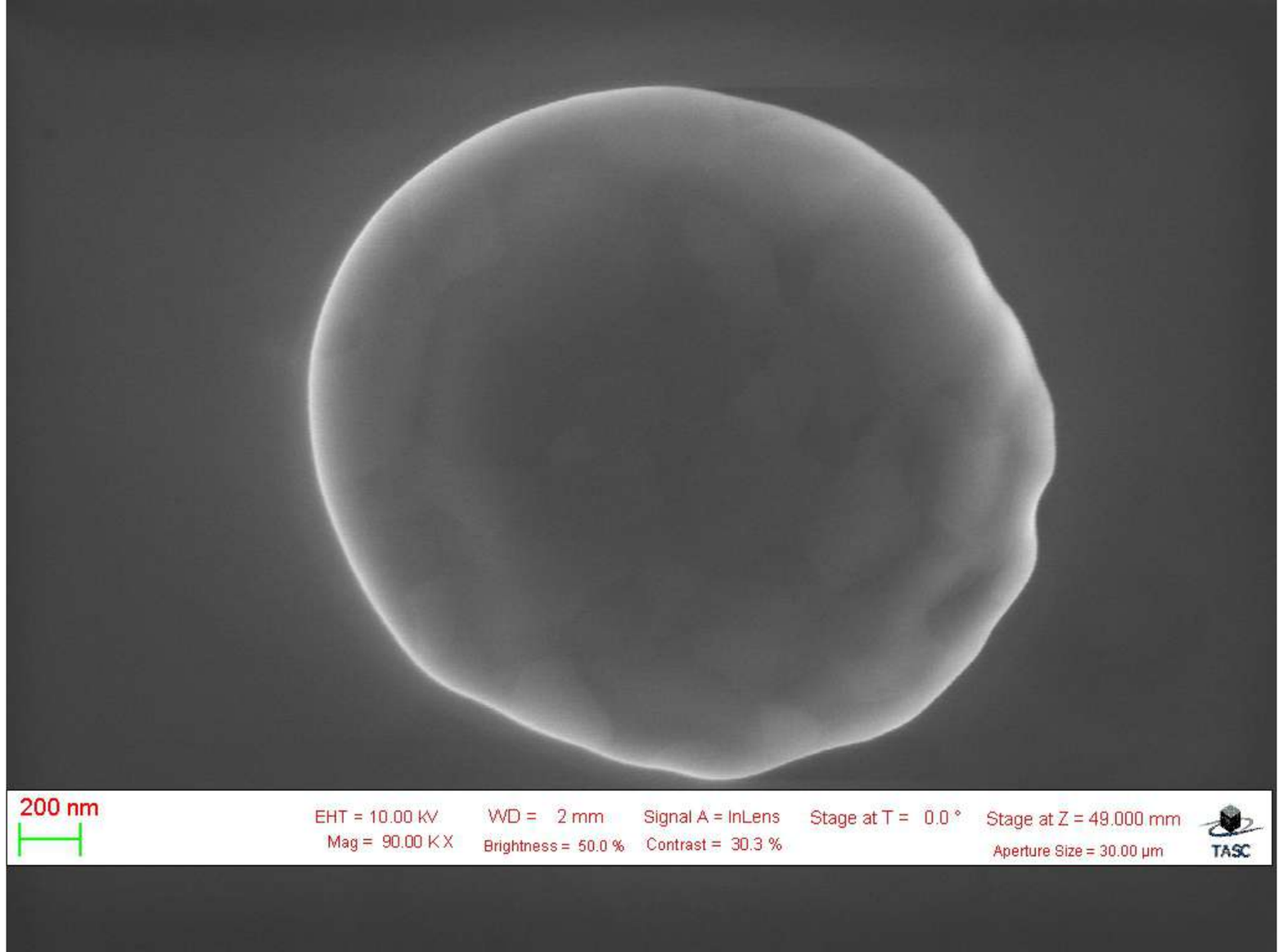}}
     \vspace{-2mm}
     \caption{The figure shows the challenges in VQA task on electron micrographs in the SEM dataset \cite{aversa2018first}.}
     \vspace{-4mm}
     \label{fig:figure2}
\end{figure}

\vspace{0mm}
Nanoimage-based VQA tasks, while advantageous, remain a significant challenge. Figure \ref{fig:figure2} illustrates the challenges in VQA tasks, which are largely attributed to high intra-class dissimilarity, high inter-class similarity, and the existence of visual patterns at multiple scales, or spatial heterogeneity. The overarching goal of this research is to develop a vision-language instruction tuning framework, utilizing pretrained LMMs such as GPT-4 for training SMMs and address the challenges in VQA tasks for enterprise adoption. The main contributions of our work are as follows:

\vspace{-1mm}
\begin{itemize}
\item The focus of our study is the development of small multimodal models (SMMs), \texttt{MVaEMa}, using visual instruction tuning. We employ GPT-4, a large, pre-trained multimodal teacher model, to generate diverse instruction-following data that better aligns with human intent. This includes the generation of detailed, context-rich question-answer pairs that explore different facets of microscopic images of nanomaterials. We utilize the high-quality, machine-generated data to provide customized instructions for training SMMs tailored to analyze electron microscopy images of nanomaterials. This teacher-student strategy enables zero-shot learning capabilities in the student models, allowing them to answer visually grounded questions without needing additional human labeling effort. Our approach facilitates knowledge distillation from proprietary LMMs to customized SMMs, improving the performance of the SMMs to be comparable to that of the LMMs on nanomaterial image analysis tasks. The pretrained SMMs can further be fine-tuned by enterprises with their in-house or proprietary data, without having to share sensitive data. 
\vspace{-4mm}
\item   We present a multimodal machine learning framework designed to process and integrate text and image data for the VQA task. It employs an image encoder with self-attention mechanism to extract salient information from images, as well as a text encoder with bidirectional self-attention to capture contextual language. The unimodal embeddings are then integrated in an image-grounded text encoder that uses cross-attention mechanism to align text representations with visual cues. This is followed by a text decoder that generates descriptive output capturing the content and context of the image, guided by various loss functions to optimize the learning process. The ultimate goal is to produce text that accurately describes or explains images to assist with interpreting microscopy images.
\end{itemize}

\vspace{-3ex}
\section{Proposed Method}
\vspace{-0.5ex}
\label{sec:method}

\vspace{-0.5mm}
\paragraph{Instruction-tuned teacher LMM:} We utilize a teacher-student strategy, employing an off-the-shelf, pre-trained large multimodal model to train small-scale multimodal model through instruction tuning on zero-shot VQA tasks. This approach accelerates the student model's learning, resulting in more accurate, relevant, and appropriate responses for tasks involving visual and linguistic information. In this work, we leverage state-of-the-art instruction-tuned foundational LMMs, such as GPT-4\cite{gpt4v}, which offers efficient and cost-effective text generation with a large context window. By utilizing this general-purpose, large-scale pre-trained vision-language model, we create instruction-following data comprising question-answer pairs by exploring various aspects, such as the microscopic image's structure and patterns, for customizing SMMs for nanomaterial image interpretation and analysis tasks. This significantly enhances their ability to autonomously handle new queries without relying on human-crafted instructions and aligns them more closely with human intentions. The GPT-4 API is accessible through Multimodal Modeling as a Service (MMaaS), an on-demand service hosted on cloud servers that accepts multimodal inputs, including both images and text, to produce outputs. This approach is similar to how Language Modeling as a Service (LMaaS) \cite{sun2022black} provides access to Large Language Models (LLMs) for language processing tasks. We generate context-augmented multimodal chain-of-thought (CoT) prompts, that consist of image captions stating the nanomaterial category, along with natural language questions as task-specific instructions, which guide GPT-4 to examine the query nanomaterial image as visual input and generate the answer to produce detailed textual descriptions in response to the natural language question. This process creates instruction-following data for training SMMs to perform VQA task, with GPT-4V leveraging its domain-specific knowledge to provide contextual descriptions based on the visual inputs and image caption, along with the query text serving as labeled data for training the SMMs.

\vspace{-3mm}
\paragraph{Multimodal Instruction-Following Data:} Using GPT-4 to generate domain-specific visual instruction tuning dataset is an effective way to train SMMs for VQA tasks related to nanomaterial images. This approach addresses the scarcity of vision-language instruction-following data and enhances SMMs domain-specific adaptation and alignment abilities, allowing them to perform comparably to proprietary LMMs without requiring excessive computational costs. Transfer learning is also used to improve generalization of SMMs, and the benefits of this approach include: (a) enhancing SMMs reasoning abilities for complex visual questions, (b) improving zero-shot learning for new questions on unseen nanoimages, (c) facilitating knowledge distillation from larger models to transfer insights about nanomaterial structures and patterns, and (d) generating diverse question-answer pairs to enrich training data and expand the smaller models capabilities. Our method employs zero-shot CoT prompting to guide GPT-4 in automatic generation of a novel instruction-following dataset (question-answer pairs) for training SMMs and involves natural language questions that analyze nanomaterials' size, distribution, morphology, and structure in microscopic images. Our approach effectively links natural language instructions (query text) with visual representations (query image), thereby enhancing SMMs' responsiveness to complex visual queries and aiding in understanding the visual representations of concept-based questions and answers. The customized CoT prompt format is as follows:

\vspace{-2mm}
\begin{tcolorbox}[colback=white!5!white,colframe=black!75!black]% [width=10cm]
\vspace{-2mm}
\textbf{Prompt 1:} **Basics** - This image depicts a nanomaterial. Identify the specific type of nanomaterial depicted in the image.? Additionally, find image scale: real-world length per unit measurement?. \textbf{Prompt 2:} **Morphology and Structure** - Describe the overall shape and morphology of the nanomaterials?. Identify any visible layers, phases, or distinct domains?. Assess consistency in size and shape, or note any variability?. \textbf{Prompt 3:} **Size and Distribution** - Estimate size/size range of nanostructures?. - Describe distribution - evenly spaced, clustered, or random?. - Comment on any aggregation or bundling visible.?. \textbf{Prompt 4:} **Surface Characteristics** - Describe surface textures - smooth, rough, distinct textures?. - Comment on any visible imperfections like defects, pores, or impurities?. \textbf{Prompt 5:} **Composition and Elements** - Note any visible evidence of compositional variations (color, brightness, contrast differences)?. -  Identify any labels or markers pointing to specific elements/compounds?. \textbf{Prompt 6:} **Interactions and Boundaries** -- Describe visual interactions: touching, fused, or separate?. - Can you distinguish boundaries between structures/phases? Or do they blend without defined borders?.
\vspace{-2mm}
\end{tcolorbox}

\vspace{-2mm}
\begin{tcolorbox}[colback=white!5!white,colframe=black!75!black]% [width=10cm]
\vspace{-2mm}
\textbf{Prompt 7:} **External Environment** - Note any visible signs of interaction between nanomaterials and surroundings (solvents, polymers, etc.)? - Identify and describe any non-nanomaterial structures/objects present?. \textbf{Prompt 8:} **Image Technique and Modifications** - Identify imaging technique used (SEM, TEM, etc.)? - Note any visible post-processing or modifications like false coloring or 3D rendering?. \textbf{Prompt 9:} **Functional Features** -  Identify any visible functional elements or regions with distinct properties?. - Note if the image shows any dynamic processes, or if it is primarily static?. \textbf{Prompt 10:}  **Context and Application** -  Identify intended use/application of nanomaterials. - Are they experimental samples or theoretical/simulation-based representations?
\vspace{-2.0mm}
\end{tcolorbox}

\vspace{-2.5ex}
\paragraph{Model Architecture:} The Figure \ref{fig:figure1} illustrates an encoder-decoder architecture designed to understand the content of visual and textual inputs, and then generate coherent, contextually appropriate responses for effectively handling complex VQA tasks that involve both visual perception and language understanding. It utilizes a visual transformer as an image encoder to split an input image into patches and convert them into a sequence of embeddings, incorporating a $\textless \textit{cls}\textgreater$ token to encapsulate the global image through self-attention mechanism. The text encoder mirrors the architecture of BERT, including a $\textless \textit{cls}\textgreater$ token at the start of the text to encapsulate the sentence's summary.  The unimodal encoders are vital to interpret a question (textual input)  about an image and then analyze the image (visual input) to provide a coherent and contextually appropriate response. The image-grounded text encoder integrates visual and textual data, focusing on key aspects via a cross-attention mechanism. It comprehends both the content and context of the image, along with the query text semantics, aiding in accurate answer generation. A $\textless \textit{Encode}\textgreater$ token appended to the text enables this multimodal integration. The output embedding of this token symbolizes the fused, multimodal image-text representation. The image-grounded text decoder employs causal attention for conditional generative decoding, signaled by a $\textless \textit{Decode}\textgreater$ token indicating the start and an end-of-sequence($\textless \textit{EOS}\textgreater$) token signaling the end of the generated text sequence; these special tokens bracket the output while guiding the auto-regressive decoding mechanism. Our proposed method for multimodal learning involves three key goals: understanding-based objectives (i.e., minimizing image-text contrastive and matching losses) for comprehending visual and textual content, generation-based objectives (i.e., minimizing language modeling loss) for producing accurate answers, and joint optimization for simultaneously training on all objectives to demonstrate exceptional proficiency in natural instruction-following and visual reasoning for the microscopic image-based VQA task. (a) \textbf{The image-text contrastive (ITC)} loss aims to minimize the distance between representations of matching image-text pairs while maximizing the distance between non-matching pairs. Minimizing ITC loss in multimodal learning, aligns matching image-text pair representations in a shared embedding space and is based on the noise-contrastive estimation principle expressed as:

\vspace{-3mm}
\resizebox{0.985\linewidth}{!}{
\begin{minipage}{\linewidth}
\begin{align*}
   L_{\text{ITC}} &= \frac{1}{2}(L_{I2T} + L_{T2I}) \\
    &= -\frac{1}{N} \sum_{i=1}^{N} \left[ \log \frac{e^{\text{sim}(v_i, t_i)/\tau}}{\sum_{j=1}^{N} e^{\text{sim}(v_i, t_j)/\tau}} + \log \frac{e^{\text{sim}(v_i, t_i)/\tau}}{\sum_{j=1}^{N} e^{\text{sim}(v_j, t_i)/\tau}} \right]
\end{align*}
 \end{minipage}
}

\vspace{1mm}
Where \( N \) is the number of image-text pairs in the batch. \( v_i \) and \( t_i \) are the embeddings of the image and text, respectively, in the \( i \)-th pair. Here, \( \text{sim}(v_i, t_i) \) is the similarity score between the \( i \)-th image embedding \( v_i \) and text embedding \( t_i \), often calculated using the dot product. \( \tau \) is the temperature parameter that scales the similarity measure. $L_{I2T}$ represents the loss for aligning images to texts(image-to-text contrastive loss) and $ L_{T2I}$ is the loss for aligning texts to images(text-to-image contrastive loss). The total ITC loss is the average of these two losses across all image-text pairs in the batch. The loss function drives  both the unimodal encoders(visual and text trasnformers) to align matching image-text pair representations and distinguish non-matching representations, fostering a cross-modal semantic understanding. (b) \textbf{The image-text matching (ITM)} loss, using binary cross-entropy loss in multimodal learning, is designed to encourage the image-grounded text encoder to correctly identify whether an image and text representation form a matching pair or not. The parameters of the image-text encoder are updated to minimize this loss, thereby improving the alignment of image-text multimodal representations in the shared embedding space. It penalizes the encoder for incorrect predictions, guiding it to learn better representations for image-text matching pairs. Let $y_i$ denote the ground truth label for the $i$-th image-text pair in a batch, where $y_i = 1$ if the image and text match (are relevant to each other), and $y_i = 0$ otherwise. Let $p_i$ be the predicted probability of pairs being positive (matched) that the $i$-th image and text match. The probability $p_i$ is computed from the output linear layer of the image-grounded text encoder by applying a sigmoid function. The binary cross-entropy loss for the ITM task over a batch of size $N$ can be formulated as follows:

\vspace{-2.5mm}
\resizebox{0.95\linewidth}{!}{
\begin{minipage}{\linewidth}
\begin{align*}
L_{\text{ITM}} = -\frac{1}{N} \sum_{i=1}^{N} \left[ y_i \log(p_i) + (1 - y_i) \log(1 - p_i) \right]
\end{align*}
 \end{minipage}
}

(c) \textbf{Language modeling loss (LM)} is particularly used for the VQA task, which focuses on generating coherent and contextually relevant text when presented with an image and a question related to that image. The image-grounded text decoder minimizes the LM loss by generating textual descriptions that accurately describe the visual content in images. Specifically, it learns to accurately predict each word in a sentence based on the preceding words and the contextual visual information provided by the corresponding image. The autoregressive decoder aims to maximize the likelihood of the correct words in the text sequence, by refining the model's ability to understand and answer questions about images. This involves minimizing the negative log-likelihood of the ground truth words under the predicted probabilities of the image-grounded text decoder, thereby leading to improved text generation that aligns with the image.

\vspace{-3mm}
\resizebox{0.925\linewidth}{!}{
\begin{minipage}{\linewidth}
\begin{align*}
L_{LM} = -\sum_{i}^{N} \log P(w_i | w_{<i}, I, Q) 
\end{align*}
 \end{minipage}
}

Where $L_{LM}$ represents the language modeling loss, $N$ is the number of words in the text, $w_i$ represents the $i$-th word in the text, $w_{<i}$ represents all words before the $i$-th word, $I$ is the image corresponding to the text, and $P(w_i | w_{<i}, I, q)$ is the probability of the $i$-th word given the preceding words and the image, as predicted by the model. $q$ refers to the question that the generated text aims to answer when conditioned on both the image $I$ and the previous words $w_{<i}$ in the sequence. During inference time, the decoder generates accurate text descriptions for a given image using the knowledge it has acquired during training.

\vspace{-3mm} 
\section{Experiments And Results}

\begin{figure}[htbp]
\centering
     \subfloat{\hspace{-0mm}\includegraphics[width=0.11\textwidth]{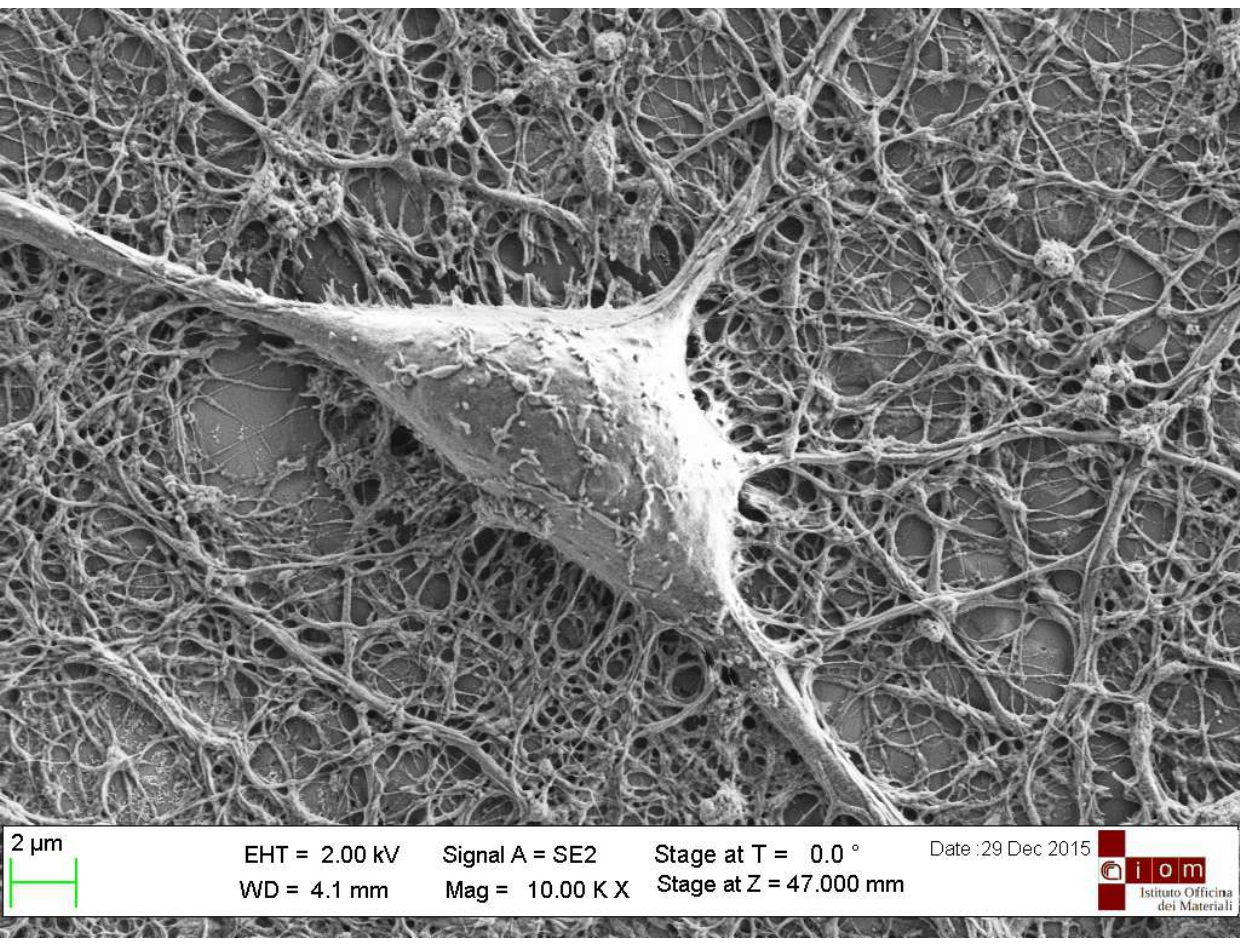}
     \includegraphics[width=0.11\textwidth]{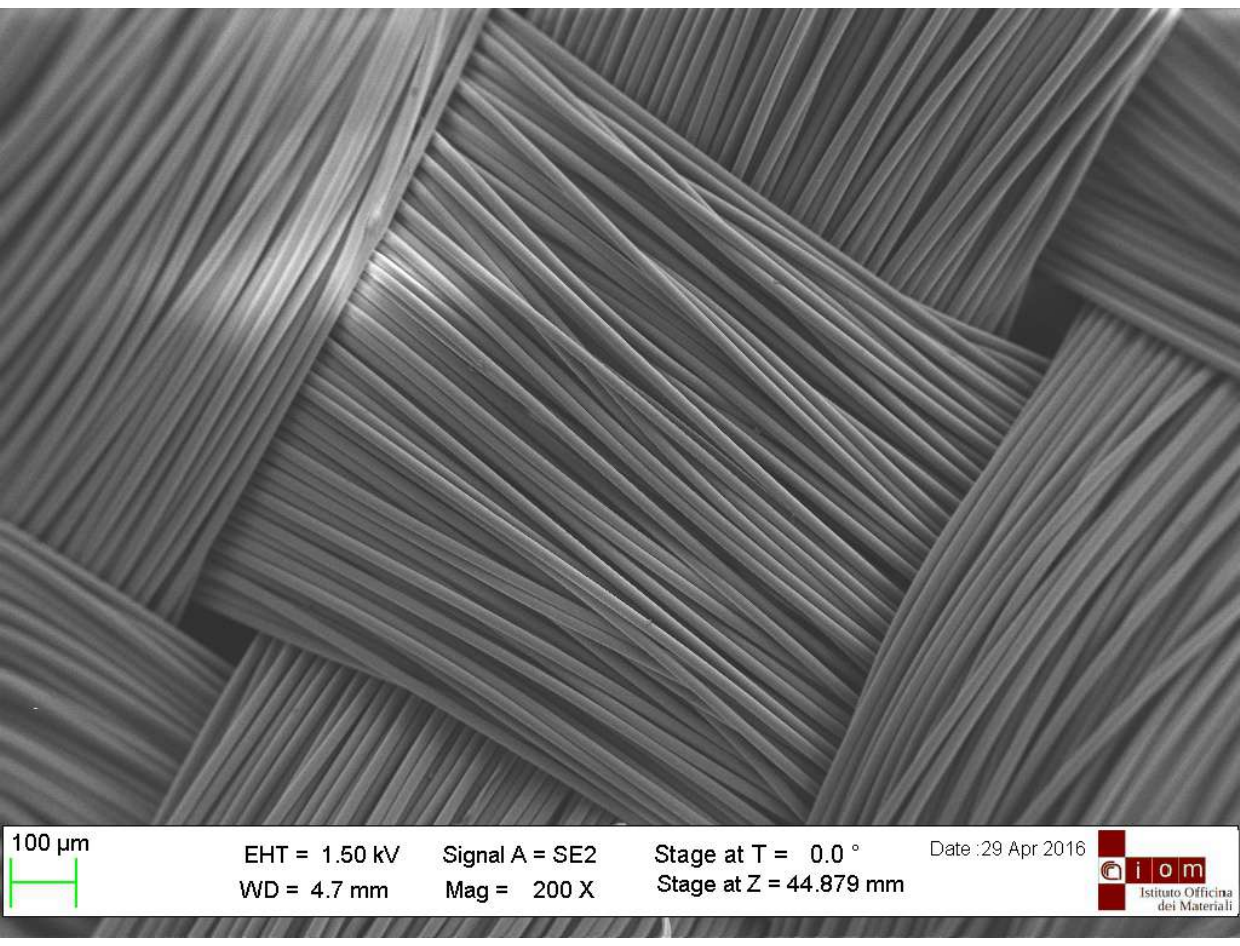}
     \includegraphics[width=0.11\textwidth]{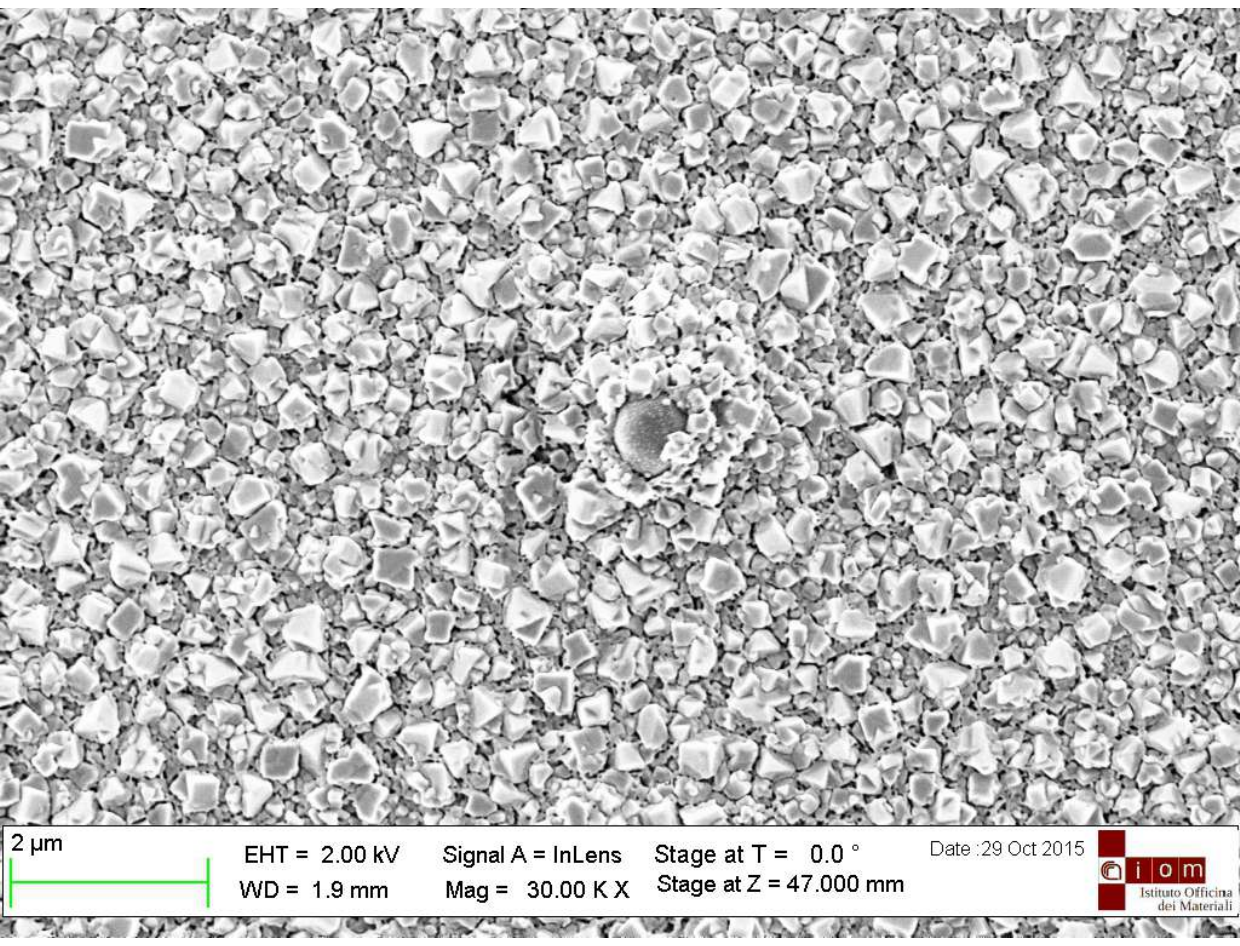}
     \includegraphics[width=0.11\textwidth]{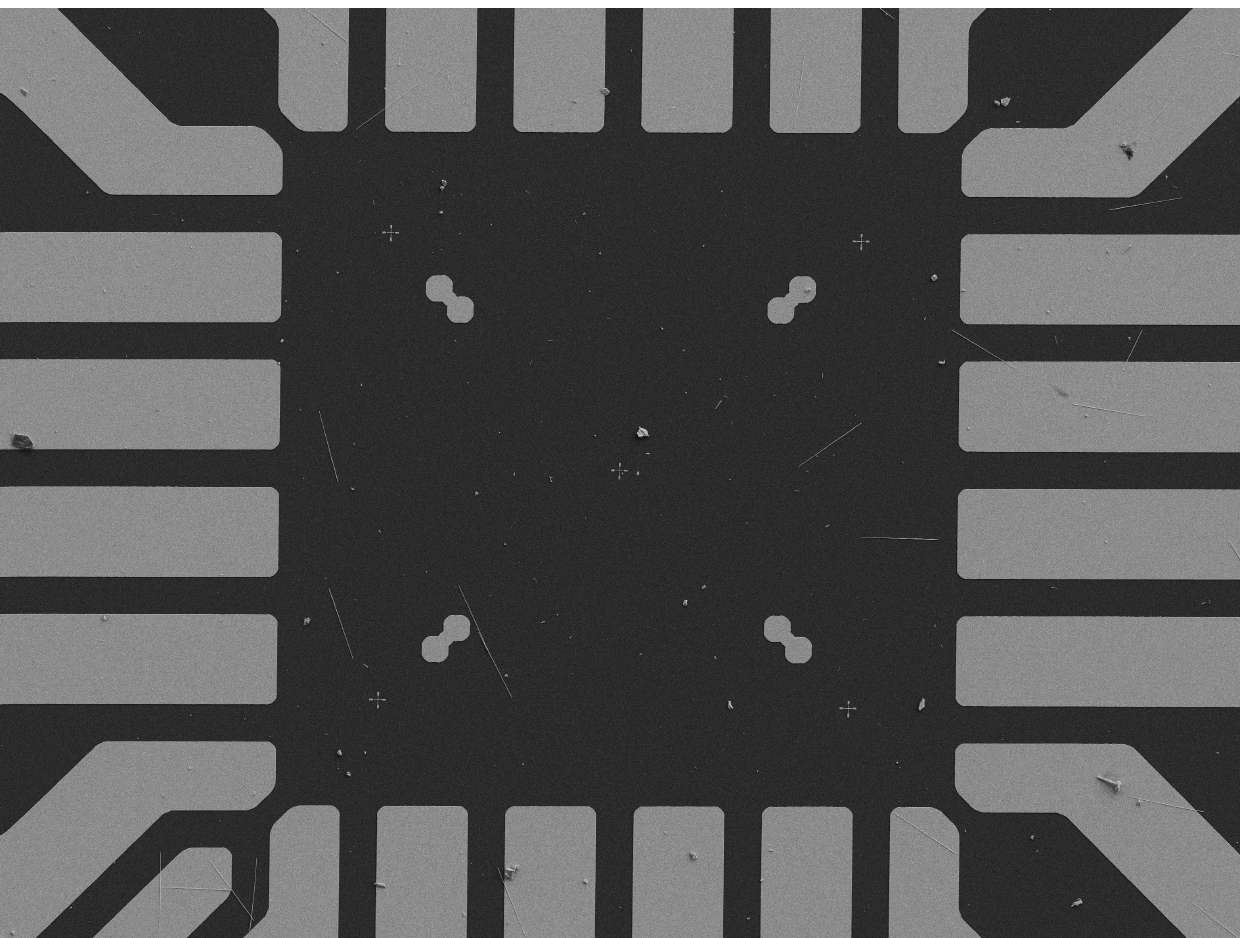}
     }
          \vspace{-12.5mm}
     \qquad
     \subfloat{\hspace{-0mm}\includegraphics[width=0.11\textwidth]{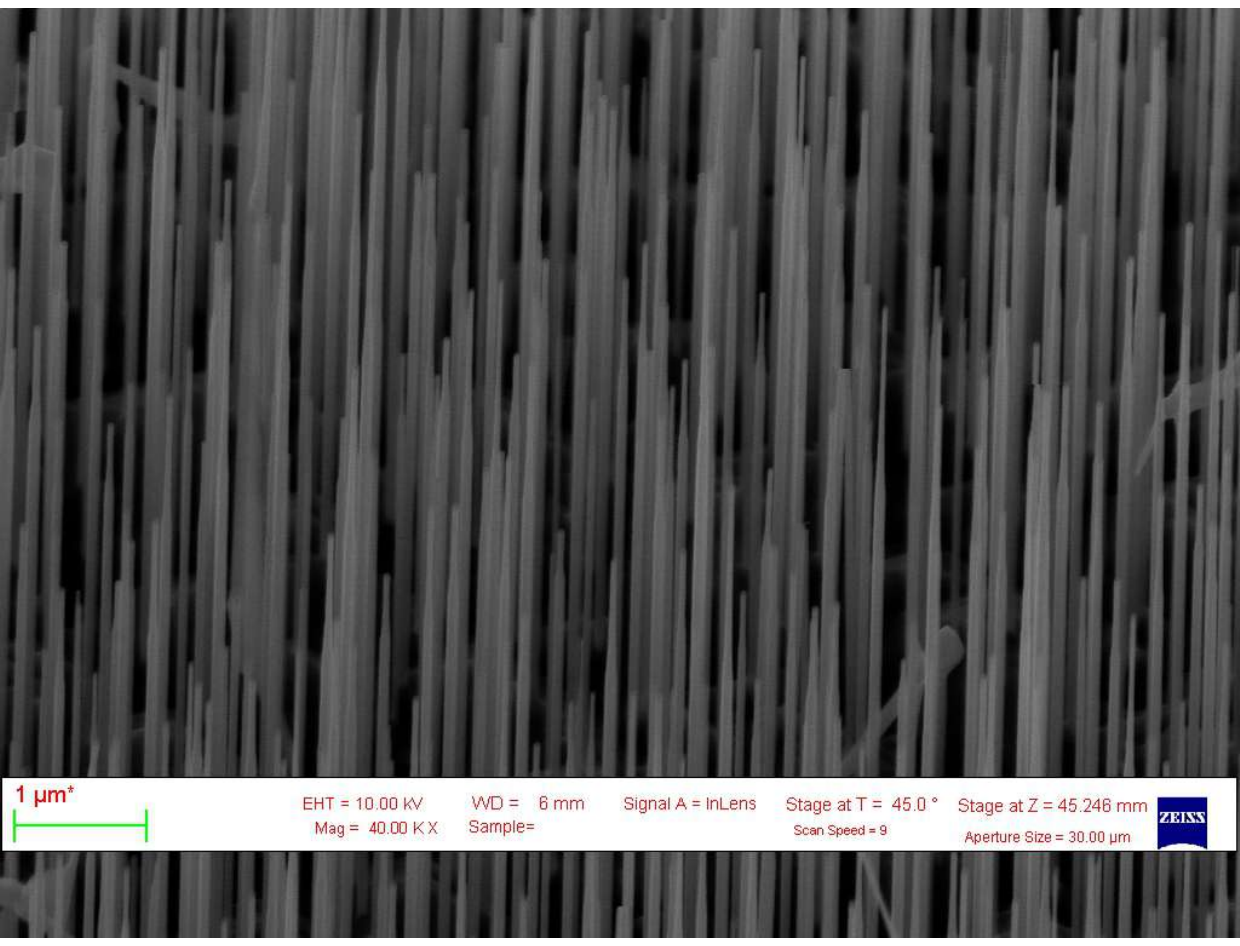}
     \includegraphics[width=0.11\textwidth]{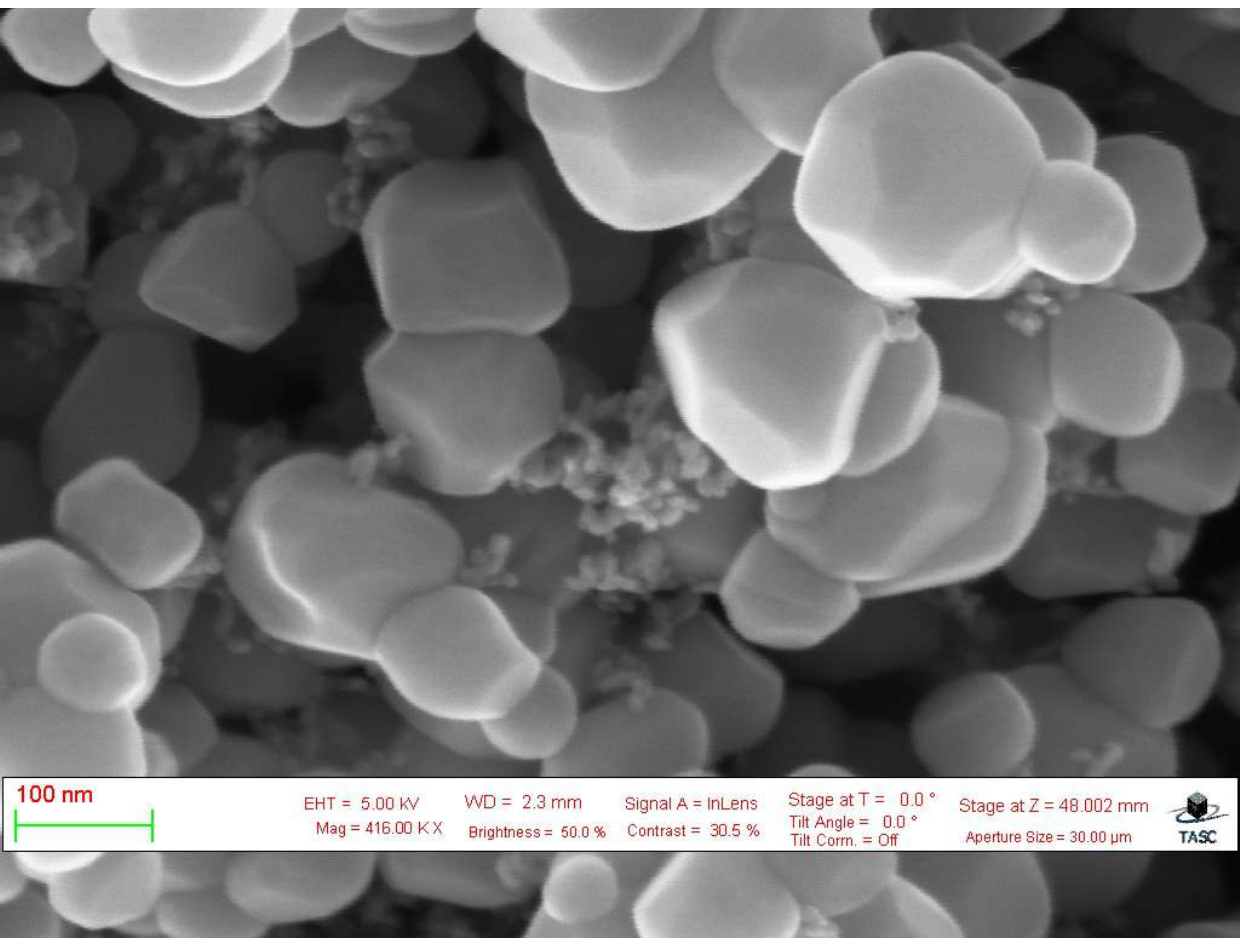}
     \includegraphics[width=0.11\textwidth]{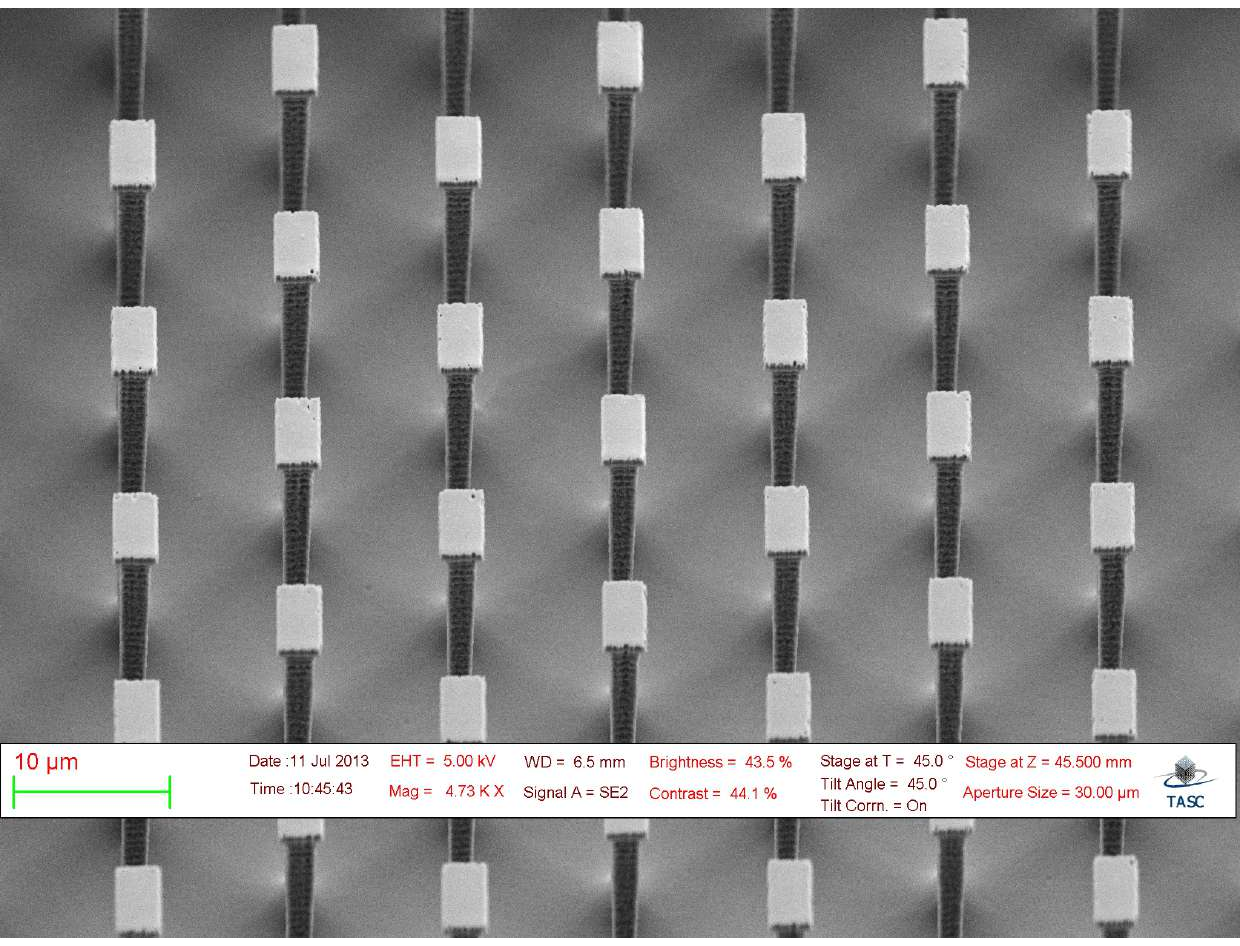}
     \includegraphics[width=0.11\textwidth]{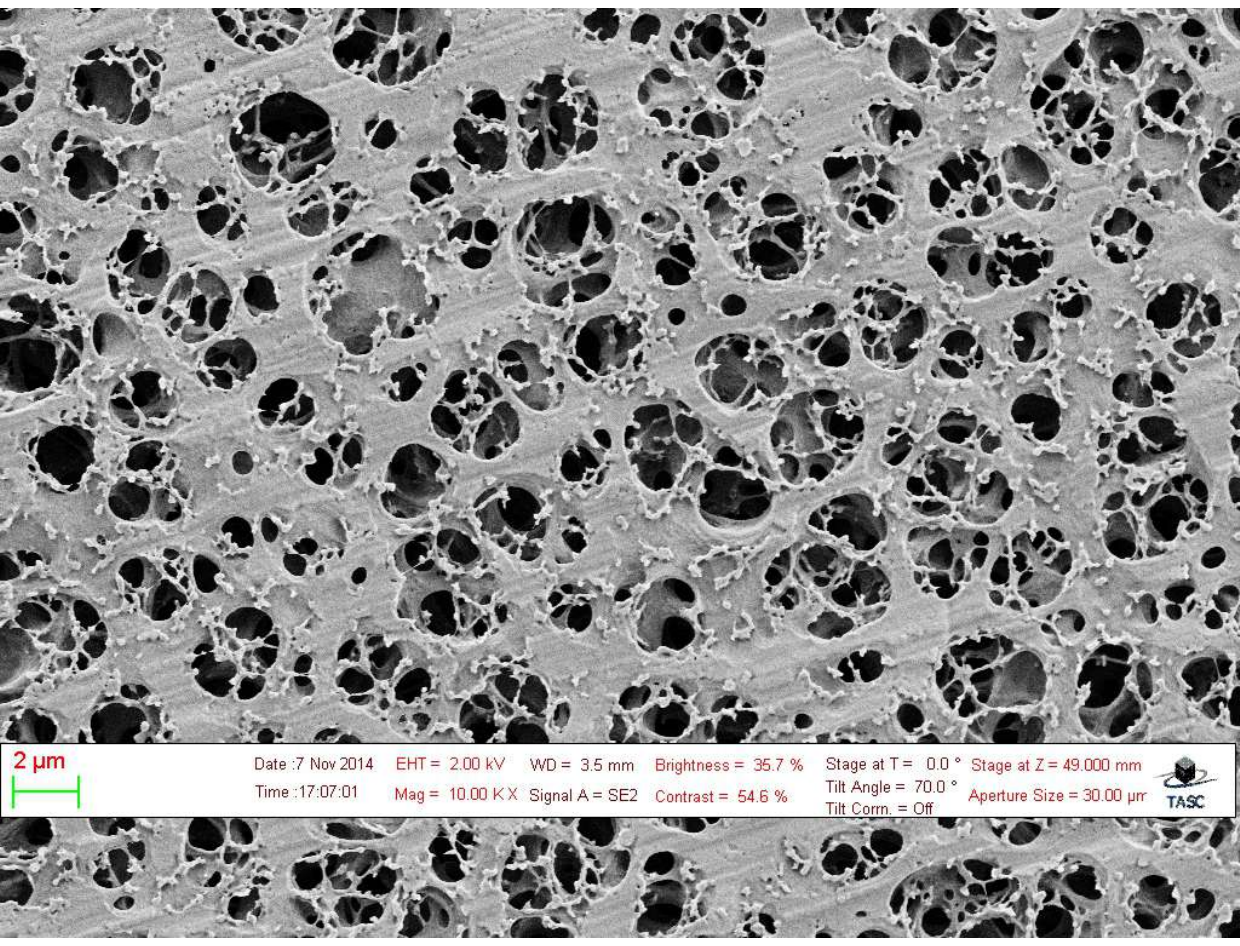}
     }
     \vspace{-13mm}
     \qquad
     \subfloat{\hspace{-0mm}\includegraphics[width=0.11\textwidth]{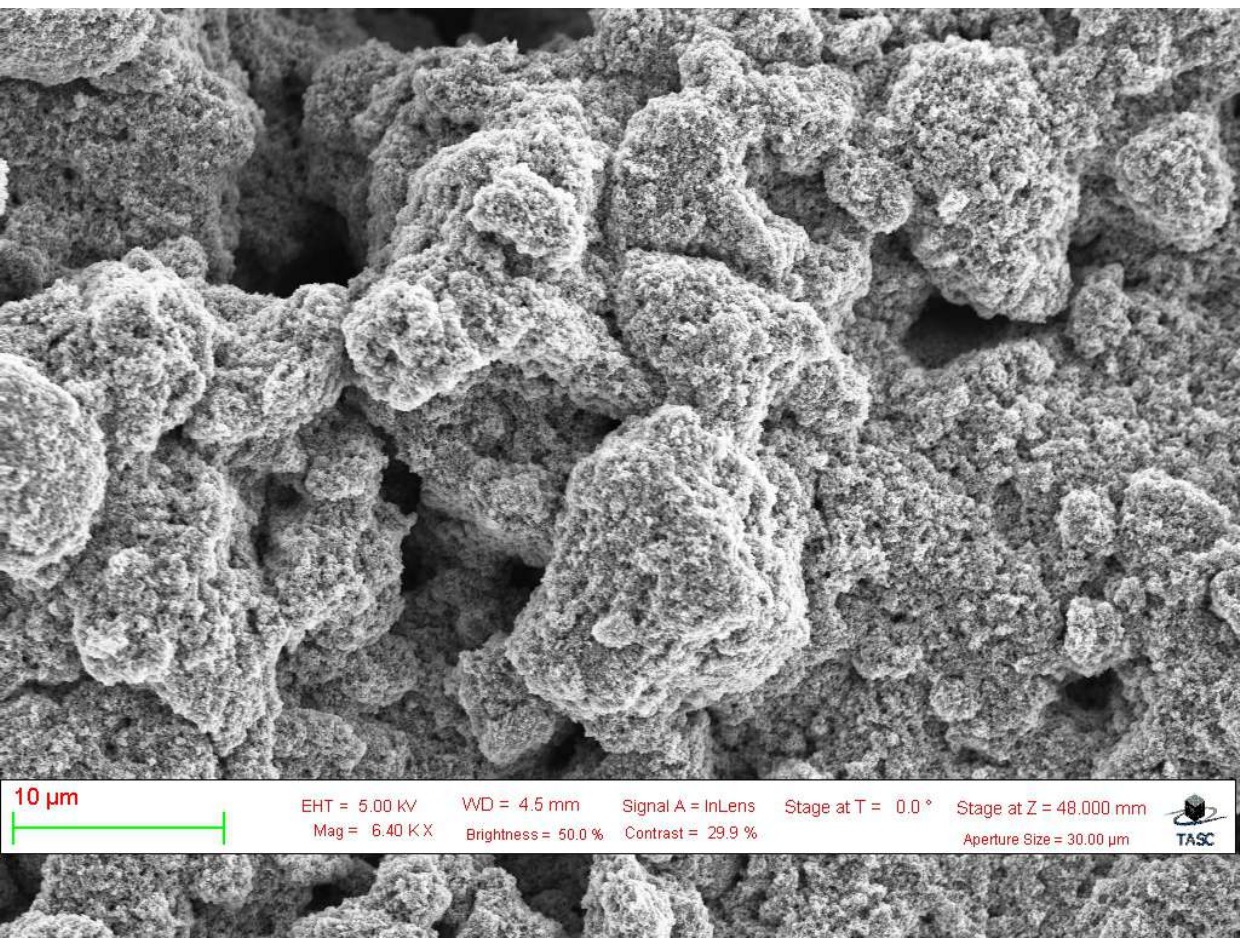}
     \includegraphics[width=0.11\textwidth]{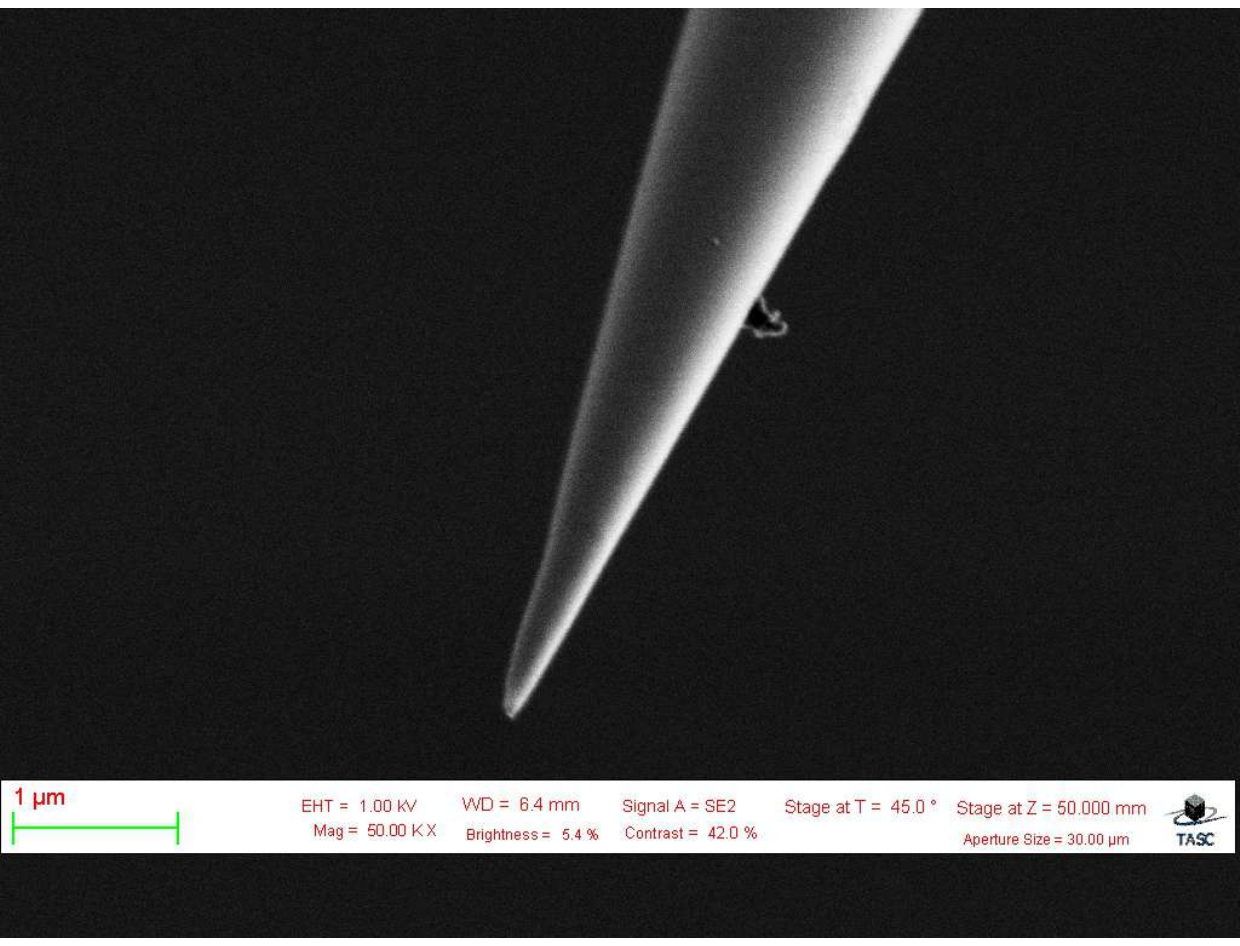}
     }
     \vspace{-9mm}
     \caption{The figure displays nanomaterials from the SEM dataset. From left to right in the first, second, and third rows, we have: \textit{biological, fibers, films, MEMS}; \textit{nanowires, particles, patterned surface, porous sponges}; and \textit{powder, tips}.}
      \vspace{0mm}
     \label{fig:illustrationpics}
\vspace{-5mm}     
\end{figure}

\vspace{-1mm}
\paragraph{Datasets:} \hspace{-2mm} Our study used the SEM dataset \cite{aversa2018first} to automate VQA task for nanomaterial image interpretation and analysis. This dataset contains 21,283 electron micrographs across 10 categories, including \textit{particles, nanowires, and patterned surfaces}. Figure \ref{fig:illustrationpics} displays the different nanomaterial categories in the SEM dataset. Initial findings \cite{modarres2017neural} on the image classification task were based on a subset, while our research utilized the complete dataset for both the zero-shot VQA and image classification tasks. In our work, to ensure a rigorous comparison with popular baseline models, we employed k-fold cross-validation, as no predefined splits were provided by the dataset curator.

\vspace{-9.5mm}

\vspace{-2mm}
\paragraph{Experimental Setup:}
The SEM dataset\cite{aversa2018first} consists of electron micrographs with dimensions of $1024\times 768\times 3$ pixels. We downscale these to $224\times 224\times 3$ pixels and normalize the micrographs by adjusting the mean and covariance to 0.5 across channels, resulting in values within [-1, 1]. We then tokenize the downscaled and normalized micrographs into non-overlapping 32 pixel patches. The patch and position embedding dimensions are set to 64. We use 10-fold cross-validation and train for 50 epochs with an initial learning rate of $1e^{-3}$ and batch size of 48. For the self, cross-modal and casual attention layers, the number of heads is 4 and key/query/value dimensionality is 16. We employ early stopping on the validation set to prevent overfitting and a learning rate scheduler that halves the learning rate if validation loss stagnates for 5 epochs. We also use the Adam optimization algorithm \cite{kingma2014adam} to update the framework's trainable parameters. We assess the performance of \texttt{MVaEMa} in instruction-following and visual reasoning capabilities using the SEM dataset on nanoimage 

\vspace{-2mm}
\begin{table*}[!ht]
\centering
\caption{The table presents the experimental results comparing the performance of the \texttt{MVaEMa} framework on the VQA task against the baseline models.}
\vspace{-3mm}
\scalebox{0.80}{
\hspace*{-5mm}\begin{tabular}{l|c|c|c|c|c|c}
\toprule
Method                                                                  & BLEU-2 ($\uparrow$)                     & BLEU-4 ($\uparrow$)                     & ROUGE-1 ($\uparrow$)                    & ROUGE-2 ($\uparrow$)                    & ROUGE-L ($\uparrow$)                   & METEOR ($\uparrow$)                     \\ \midrule
\begin{tabular}[c]{@{}l@{}} InstructBLIP\cite{dai2305instructblip} \end{tabular}            & 0.570$\pm$0.063          & 0.457$\pm$0.078          & 0.745$\pm$0.032          & 0.648$\pm$0.011          & 0.705$\pm$0.042          & 0.738$\pm$0.048          \\ \midrule
\begin{tabular}[c]{@{}l@{}}LLaVA\cite{liu2023visual} \end{tabular}         & 0.620$\pm$0.070          & 0.512$\pm$0.085          & 0.760$\pm$0.032          & 0.668$\pm$0.011          & 0.723$\pm$0.042          & 0.753$\pm$0.046          \\ \midrule
\begin{tabular}[c]{@{}l@{}}MiniGPT-4\cite{zhu2023minigpt} \end{tabular}     & 0.680$\pm$0.075          & 0.572$\pm$0.090          & 0.790$\pm$0.033          & 0.698$\pm$0.012          & 0.753$\pm$0.043          & 0.783$\pm$0.047          \\ \midrule
\begin{tabular}[c]{@{}l@{}}\textbf{MVaEMa} \end{tabular} & \textbf{0.780 $\pm$0.085} & \textbf{0.709 $\pm$0.105} & \textbf{0.860 $\pm$0.036}       &     \textbf{0.765 $\pm$0.014}     & \textbf{0.822 $\pm$0.050} &      \textbf{0.853 $\pm$0.055}     \\  \bottomrule
\end{tabular}
}
\label{captioning_results1}
\vspace{-4mm}
\end{table*}

\begin{table}[ht!]
\footnotesize
\centering
\setlength{\tabcolsep}{5pt}
\caption{The table compares our method to baseline algorithms, including vision-based supervised convolutional neural networks (ConvNets), vision transformers (ViTs), and self-supervised learning (VSL) algorithms on nanomaterial image classification task.}
\vspace{-3mm}
\label{tab:table2}
\begin{tabular}{c|c|c|c}
\toprule
\multicolumn{2}{c|}{\textbf{Algorithms}} & \textbf{Top-1} & \textbf{Top-5} \\ 
\midrule
\multirow{6}{*}{\rotatebox[origin=c]{90}{\textbf{ConvNets}}} 
& AlexNet(\cite{krizhevsky2017imagenet}) & 0.528 & 0.827 \\
& DenseNet(\cite{huang2017densely}) & 0.569 & 0.929 \\
& ResNet(\cite{he2016deep}) & 0.485 & 0.897 \\
& VGG(\cite{simonyan2014very}) & 0.538 & 0.808 \\
& GoogleNet(\cite{szegedy2015going}) & 0.609 & 0.969 \\
& SqueezeNet(\cite{iandola2016squeezenet}) & 0.404 & 0.698 \\
\midrule
\multirow{6}{*}{\rotatebox[origin=c]{90}{\textbf{VSL}}} 
& Barlowtwins\cite{zbontar2021barlow} & 0.148 & 0.410 \\
& SimCLR\cite{chen2020simple} & 0.130 & 0.379 \\
& byol\cite{grill2020bootstrap} & 0.143 & 0.453 \\
& moco\cite{he2020momentum} & 0.169 & 0.472 \\
& nnclr\cite{dwibedi2021little} & 0.158 & 0.563 \\
& simsiam\cite{chen2021exploring} & 0.188 & 0.535 \\
\midrule
\multirow{24}{*}{\rotatebox[origin=c]{90}{\textbf{Vision Transformers (ViTs)}}} 
& CCT\cite{hassani2021escaping} & 0.570 & 0.981 \\
& CVT\cite{CVT} & 0.577 & 0.930 \\
& ConViT\cite{ConViT} & 0.609 & 0.957 \\
& ConvVT\cite{CVT} & 0.319 & 0.921 \\
& CrossViT\cite{Crossvit} & 0.442 & 0.915 \\
& PVTC\cite{PVT} & 0.596 & 0.964 \\
& SwinT\cite{SwinT} & 0.707 & 0.993 \\
& VanillaViT\cite{dosovitskiy2020image} & 0.655 & 0.970 \\
& Visformer\cite{visformer} & 0.398 & 0.856 \\
& ATS\cite{fayyaz2021ats} & 0.540 & 0.973 \\
& CaiT\cite{CaiT} & 0.657 & 0.989 \\
& DeepViT\cite{Deepvit} & 0.546 & 0.988 \\
& Dino\cite{Dino} & 0.049 & 0.437 \\
& Distillation\cite{Distillation} & 0.533 & 0.955 \\
& LeViT\cite{Levit} & 0.624 & 0.970 \\
& MA\cite{MA} & 0.202 & 0.491 \\
& NesT\cite{Nest} & 0.660 & 0.985 \\
& PatchMerger\cite{PatchMerger} & 0.578 & 0.975 \\
& PiT\cite{PiT} & 0.555 & 0.979 \\
& RegionViT\cite{Regionvit} & 0.606 & 0.948 \\
& SMIM\cite{SMIM} & 0.171 & 0.646 \\
& T2TViT\cite{T2TViT} & 0.749 & 0.992 \\
& ViT-SD\cite{ViT-SD} & 0.597 & 0.973 \\
\midrule
& \textbf{MVaEMa} & \textbf{0.947} & \textbf{0.988} \\
\bottomrule
\end{tabular}
\vspace{-8mm}
\end{table}

\newpage
analysis tasks. In our work, we utilize GPT-4 to obtain a multimodal instruction-following dataset(question-answer pairs) on the SEM dataset. We set the temperature to 0.25 to control randomness in text generation and top-p sampling to 0.1 to narrow down word choices for more deterministic output. Additionally, we set the maximum number of output tokens to 3500. We implement the framework in pytorch\cite{paszke2019pytorch} and pretrained on 4 $\times$ V100 GPUs. Due to the potentially high computational cost of using prompting with large multimodal models, we conducted each experiment twice and reported the averaged results.

\vspace{-2mm}
\begin{table}[ht!]
\footnotesize
\centering
\setlength{\tabcolsep}{4pt}
\caption{The table presents a performance comparison of supervised-learning GNNs, self-supervised GCL algorithms, and our novel method for nanomaterial classification task.}
\label{tab:table3}
\vspace{-3mm}
\begin{tabular}{cc|c|c}
\hline
\multicolumn{2}{c|}{\textbf{Algorithms}} & \textbf{Top-1} & \textbf{Top-5}  \\ \hline
\multicolumn{1}{c|}{\multirow{4}{*}{\rotatebox[origin=c]{90}{\textbf{GCL}}}} 
& GBT\cite{bielak2021graph} & 0.547 & 0.706 \\
\multicolumn{1}{c|}{} & GRACE\cite{zhu2020deep} & 0.598 & 0.750 \\
\multicolumn{1}{c|}{} & BGRL\cite{thakoor2021bootstrapped} & 0.556 & 0.696 \\
\multicolumn{1}{c|}{} & InfoGraph\cite{sun2019infograph} & 0.526 & 0.702 \\
\hline
\multicolumn{1}{c|}{\multirow{15}{*}{\rotatebox[origin=c]{90}{\textbf{Graph Neural Networks}}}} 
& APPNP\cite{klicpera2018predict} & 0.632 & 0.786 \\
\multicolumn{1}{c|}{} & AGNN\cite{thekumparampil2018attention} & 0.538 & 0.894 \\
\multicolumn{1}{c|}{} & ARMA\cite{bianchi2021graph} & 0.582 & 0.987 \\
\multicolumn{1}{c|}{} & DNA\cite{fey2019just} & 0.622 & 0.916 \\
\multicolumn{1}{c|}{} & GAT\cite{velivckovic2017graph} & 0.491 & 0.985 \\
\multicolumn{1}{c|}{} & GGConv\cite{li2015gated} & 0.563 & 0.992 \\
\multicolumn{1}{c|}{} & GraphConv\cite{morris2019weisfeiler} & 0.658 & 0.996 \\
\multicolumn{1}{c|}{} & GCN2Conv\cite{chen} & 0.732 & 0.998 \\
\multicolumn{1}{c|}{} & ChebConv\cite{defferrard2016convolutional} & 0.504 & 0.951 \\ 
\multicolumn{1}{c|}{} & GraphConv\cite{morris2019weisfeiler} & 0.509 & 0.993 \\
\multicolumn{1}{c|}{} & GraphUNet\cite{gao2019graph} & 0.657 & 0.978 \\
\multicolumn{1}{c|}{} & MPNN\cite{gilmer2017neural} & 0.603 & 0.999 \\
\multicolumn{1}{c|}{} & RGGConv\cite{bresson2017residual} & 0.618 & 0.961 \\
\multicolumn{1}{c|}{} & SuperGAT\cite{kim2022find} & 0.598 & 0.985 \\
\multicolumn{1}{c|}{} & TAGConv\cite{du2017topology} & 0.598 & 0.999 \\
\hline
\multicolumn{1}{c|}{} & \textbf{MVaEMa} & \textbf{0.947} & \textbf{0.988} \\  \bottomrule
\end{tabular}
\vspace{-3mm}
\end{table}

\vspace{-2mm}
\paragraph{VQA Results:}
In VQA tasks, text quality is evaluated using metrics such as BLEU, METEOR, and ROUGE. BLEU-N measures the similarity of machine-generated text to reference texts by analyzing overlapping n-word phrases, focusing on precision but not fluency or grammar. METEOR focuses on the harmonic mean of unigram precision and recall, and incorporates linguistic concepts like stemming and synonym matching for effective paraphrase handling, comparing generated text with reference text. ROUGE-N is an n-gram recall metric that computes overlapping n-grams between candidate and reference texts to evaluate completeness of generated answers in VQA task, with variants such as ROUGE-L measuring longest common subsequence matches. Finally, these metrics emphasize different aspects of long-form text generation, such as similarity, linguistic quality, and coherence. In comparison to other concurrent multimodal models that demonstrate competence in generating long-form responses like InstructBLIP\cite{dai2305instructblip}, LLaVA\cite{liu2023visual}, and MiniGPT-4\cite{zhu2023minigpt}, the \texttt{MVaEMa} framework stands out in its ability to generate long-form responses that seamlessly integrate fine-grained visual details with logically coherent reasoning - a hallmark often absent in other long-form generative models. We argue the preference for long or short responses in VQA task is not absolute, but should consider the question requirements, user needs and preferences, and context of use. The goal is balancing sufficient information with clarity and conciseness. Table \ref{captioning_results1} reports the experimental results on the VQA task in comparision to the baselines. Unlike LLaVA and MiniGPT-4, which produce lengthy yet less relevant responses, the \texttt{MVaEMa} framework adaptively adjusts the response length to optimize relevance. These advantages arise from the diverse instruction tuning data and effective architectural design of \texttt{MVaEMa} framework. For comparison with our algorithm, we fine-tuned the baselines on the nanoimage analysis tasks and evaluated the performance of the baselines.

\vspace{-3mm}
\begin{table}[htbp]
\footnotesize
\centering
\resizebox{0.5\textwidth}{!}{%
\begin{tabular}{@{}c|ccc|c@{}}
\toprule
\multirow{2}{*}{\textbf{Category}}   & \multicolumn{3}{c|}{\textbf{Multi-class metrics}}                                    \\ \cmidrule(lr){2-4}
                            & \multicolumn{1}{c|}{\textbf{Precision}} & \multicolumn{1}{c|}{\textbf{Recall}} & \textbf{F1 Score} &                        \\ \midrule
Biological                  & \multicolumn{1}{c|}{0.959$\pm$0.009}     & \multicolumn{1}{c|}{0.975$\pm$0.007}      & 0.965$\pm$0.013 \\
Tips                        & \multicolumn{1}{c|}{0.937$\pm$0.005}     & \multicolumn{1}{c|}{0.949$\pm$0.008}      &  0.946$\pm$0.011                           \\
Fibres                      & \multicolumn{1}{c|}{0.983$\pm$0.007}     & \multicolumn{1}{c|}{0.992$\pm$0.012}      &  0.990$\pm$0.014                        \\
Porous Sponge               & \multicolumn{1}{c|}{0.957$\pm$0.014}     & \multicolumn{1}{c|}{0.969$\pm$0.013}      &  0.953$\pm$0.010                            \\
Films Coated Surface        & \multicolumn{1}{c|}{0.967$\pm$0.005}     & \multicolumn{1}{c|}{0.963$\pm$0.009}      &  0.971$\pm$0.008                  \\
Patterned surface           & \multicolumn{1}{c|}{0.975$\pm$0.016}     & \multicolumn{1}{c|}{0.971$\pm$0.006}      &  0.970$\pm$0.014                         \\
Nanowires                   & \multicolumn{1}{c|}{0.967$\pm$0.012}     & \multicolumn{1}{c|}{0.974$\pm$0.007}      &  0.977$\pm$0.011                        \\
Particles                   & \multicolumn{1}{c|}{0.963$\pm$0.006}     & \multicolumn{1}{c|}{0.965$\pm$0.011}      &  0.957$\pm$0.023                       \\
MEMS devices                & \multicolumn{1}{c|}{0.967$\pm$0.011}     & \multicolumn{1}{c|}{0.960$\pm$0.008}      &  0.951$\pm$0.009                        \\
Powder                      & \multicolumn{1}{c|}{0.969$\pm$0.014}     & \multicolumn{1}{c|}{0.956$\pm$0.009}      &  0.945$\pm$0.011
                          \\ \bottomrule
\end{tabular}}
\vspace{-3mm}
\caption{The table shows the effectiveness of our proposed framework in terms of precision, recall, and F1-score for accurately classifying nanomaterials of different categories.}
\label{tab:table4}
\vspace{-4mm}
\end{table}

\vspace{-2mm}
\paragraph{Image Classification Results:} We evaluated our proposed framework against commonly used computer vision baselines like ConvNets, ViTs\cite{philvformer, neelayvformer}, and self-supervised vision contrastive learning (VCL)\cite{susmelj2020lightly} algorithms on the zero-shot image classification task. In this setting, the multimodal prompt (query image and query text) did not contain the image caption as shown in Figure \ref{fig:figure1}. Additionally, we compared the framework's performance to supervised graph neural networks (GNNs\cite{rozemberczki2021pytorch, Fey/Lenssen/2019}), and graph contrastive learning (GCL\cite{Zhu:2021tu}) algorithms. We compared algorithms measuring Top-N accuracy for N = 1, 5. Tables \ref{tab:table2} and \ref{tab:table3} present experimental findings that compare the performance of our proposed framework against various baseline algorithms. Under consistent experimental settings, our framework exceeded the baselines, demonstrating $26.44\%$ relative improvement in Top-1 accuracy compared to the next-best model, T2TViT \cite{T2TViT}. 
We conducted additional experiments to evaluate the performance of our proposed framework for categorizing electron micrographs across diverse nanomaterial categories characterized by varied structures, patterns and complexity. To enable a comprehensive evaluation, we adopted a multi-metric approach that utilized a confusion matrix encompassing true positives, false negatives, true negatives, and false positives to compute precision, recall, and F1-score metrics. The confusion matrix provides insights into how our framework categorized micrographs across different nanomaterial types. The results in Table \ref{tab:table4} demonstrate that our framework could generalize across various nanomaterial categories, even those with complex patterns.

\vspace{-1mm}
\paragraph{Ablation Study:} To validate the effectiveness of the methods in our framework, we conducted ablation studies by systematically disabling certain methods to create ablated variants and were evaluated using the SEM dataset \cite{aversa2018first}, with our original framework as the baseline for comparison on both VQA and image classification tasks. The ablation study enables us to verify the efficacy of our methods, substantiate their neural network designs, and justify their inclusion in the framework. A substantial performance decrease in the ablated variants compared to the baseline highlights the importance of the omitted methods. We evaluate the ablated variants performance on metrics such as precision and recall for image classification tasks, or other relevant measures for VQA task. The ablated variants that exclude the image-text contrastive loss(ITC), binary cross entropy loss (CTC), and the self-attention(SA), cross attention(CA), causal self-attention(CSA) mechanisms are denoted as proposed framework ``w/o ITC", ``w/o CTC", ``w/o SA", ``w/o CA", and ``w/o CSA" respectively. The abbreviation ``w/o" stands for ``without".  Across all ablated variants, we observe a consistent decline in performance metrics compared to the baseline. These results clearly validate the crucial contribution of each omitted method through our ablation studies. Tables \ref{tab:ablation1} and \ref{tab:ablation2} shows the ablation study results on the VQA and classification tasks, respectively. The percentage ($\%$) drop is reported compared to the baseline and signifies the importance of the disabled method in the original framework.

\vspace{-3mm}
\begin{table}[!ht]
\centering
\caption{Our ablation study disables individual methods to assess their contributions, consistently revealing their significance through performance declines in ablated variants compared to the baseline. The table shows the ablation study results on the VQA task.}
\vspace{-3mm}
\scalebox{0.80}{
\hspace{-5mm}\begin{tabular}{l|c|c|c}
\toprule
Method                                                      & BLEU-4 ($\uparrow$) & ROUGE-L ($\uparrow$) & METEOR ($\uparrow$) \\ \midrule
\begin{tabular}[c]{@{}l@{}} \textbf{MVaEMa w/o ITC} \end{tabular}        & 0.579 (\(\downarrow\) 18.34\%)     & 0.670 (\(\downarrow\) 18.49\%)     & 0.696 (\(\downarrow\) 18.41\%)     \\ \midrule
\begin{tabular}[c]{@{}l@{}} \textbf{MVaEMa w/o CTC} \end{tabular}     & 0.569 (\(\downarrow\) 19.75\%)     & 0.670 (\(\downarrow\) 18.49\%)     & 0.696 (\(\downarrow\) 18.41\%)     \\ \midrule
\begin{tabular}[c]{@{}l@{}} \textbf{MVaEMa w/o SA} \end{tabular} & 0.682 (\(\downarrow\) 3.81\%)     & 0.775 (\(\downarrow\) 5.72\%)     & 0.794 (\(\downarrow\) 6.92\%)     \\ \midrule
\begin{tabular}[c]{@{}l@{}} \textbf{MVaEMa w/o CA} \end{tabular} & 0.652 (\(\downarrow\) 8.04\%)     & 0.755 (\(\downarrow\) 8.15\%)     & 0.769 (\(\downarrow\) 9.85\%)     \\ \midrule
\begin{tabular}[c]{@{}l@{}} \textbf{MVaEMa w/o CSA} \end{tabular}            & 0.649 (\(\downarrow\) 8.46\%)     & 0.740 (\(\downarrow\) 9.98\%)     & 0.761 (\(\downarrow\) 10.79\%)  \\ \midrule
\begin{tabular}[c]{@{}l@{}}\textbf{Baseline (MVaEMa)} \end{tabular}            & \textbf{0.709} & \textbf{0.822} & \textbf{0.853} \\ \bottomrule
\end{tabular}
}
\label{tab:ablation1}
\vspace{-4mm}
\end{table}

\vspace{-3mm}
\begin{table}[htbp]
\footnotesize
\centering
\caption{Ablation study results on image classification task.}
\vspace{-2mm}
\setlength{\tabcolsep}{3.5pt}
\resizebox{0.525\textwidth}{!}{%
\hspace{-5mm}\begin{tabular}{cc|c|c|c|}
\hline
\multicolumn{2}{c|}{\textbf{Algorithms}} & \textbf{Avg-Precision} & \textbf{Avg-Recall} & \textbf{Avg-F1 Score} \\ \toprule
\multicolumn{1}{c}{\multirow{4}{*}{\rotatebox[origin=c]{90}{\textbf{}}}} & \textbf{MVaEMa w/o ITC} & 0.752 (↓21.99\%) & 0.731 (↓24.41\%) & 0.717 (↓25.47\%) \\ \midrule
\multicolumn{1}{c}{} & \textbf{MVaEMa w/o CTC} & 0.746 (↓22.61\%) & 0.773 (↓20.06\%) & 0.759 (↓21.10\%)  \\ \midrule
\multicolumn{1}{c}{} & \textbf{MVaEMa w/o SA} & 0.927 (↓3.84\%) & 0.912 (↓5.69\%) & 0.895 (↓6.96\%) \\ \midrule
\multicolumn{1}{c}{} & \textbf{MVaEMa w/o CA} & 0.867 (↓10.06\%) & 0.872 (↓9.82\%) & 0.860 (↓10.60\%) \\ \midrule
\multicolumn{1}{c}{} & \textbf{MVaEMa w/o CSA} & 0.843 (↓12.55\%) & 0.866 (↓10.44\%) & 0.885 (↓8.00\%) \\ \midrule
\multicolumn{1}{c}{} & \textbf{MVaEMa} & 0.964 & 0.967 & 0.962 \\ \hline
\end{tabular}}
\label{tab:ablation2}
\vspace{-3mm}
\end{table}

\vspace{-4mm}
\section{Conclusion}
\vspace{-1mm}
In summary, our research introduces a small-scale instruct-tuned language-and-vision assistant for electron micrograph analysis customized through a novel instruction-following multimodal dataset generated by GPT-4 Turbo with vision. The proposed framework performs VQA tasks, particularly in nanomaterial image analysis tasks, and demonstrates potential in interpreting complex visual and textual questions. The approach enables more efficient and secure enterprise applications, as the pretrained framework can be fine-tuned with proprietary data without external data exposure.

\begin{table*}[!htb]
    \caption{The table shows illustrative microscopic images, ground-truth and model-generated answers for the question to describe the overall shape and morphology of the nanomaterials. In addition, we report the BLUE-2, ROGUE-L, METEOR scores.}
      \centering 
      \vspace{-2mm}
        \begin{tabular}{|>{\centering\arraybackslash}m{2cm}|m{5.0cm}|m{5.0cm}|m{1.22cm}|m{1.53cm}|m{1.3cm}|}
        \hline 
        Image & Ground Truth & Answers & BLUE-2 & ROGUE-L & METEOR \\ \hline
        \includegraphics[width=2cm,height=1.5cm,keepaspectratio]{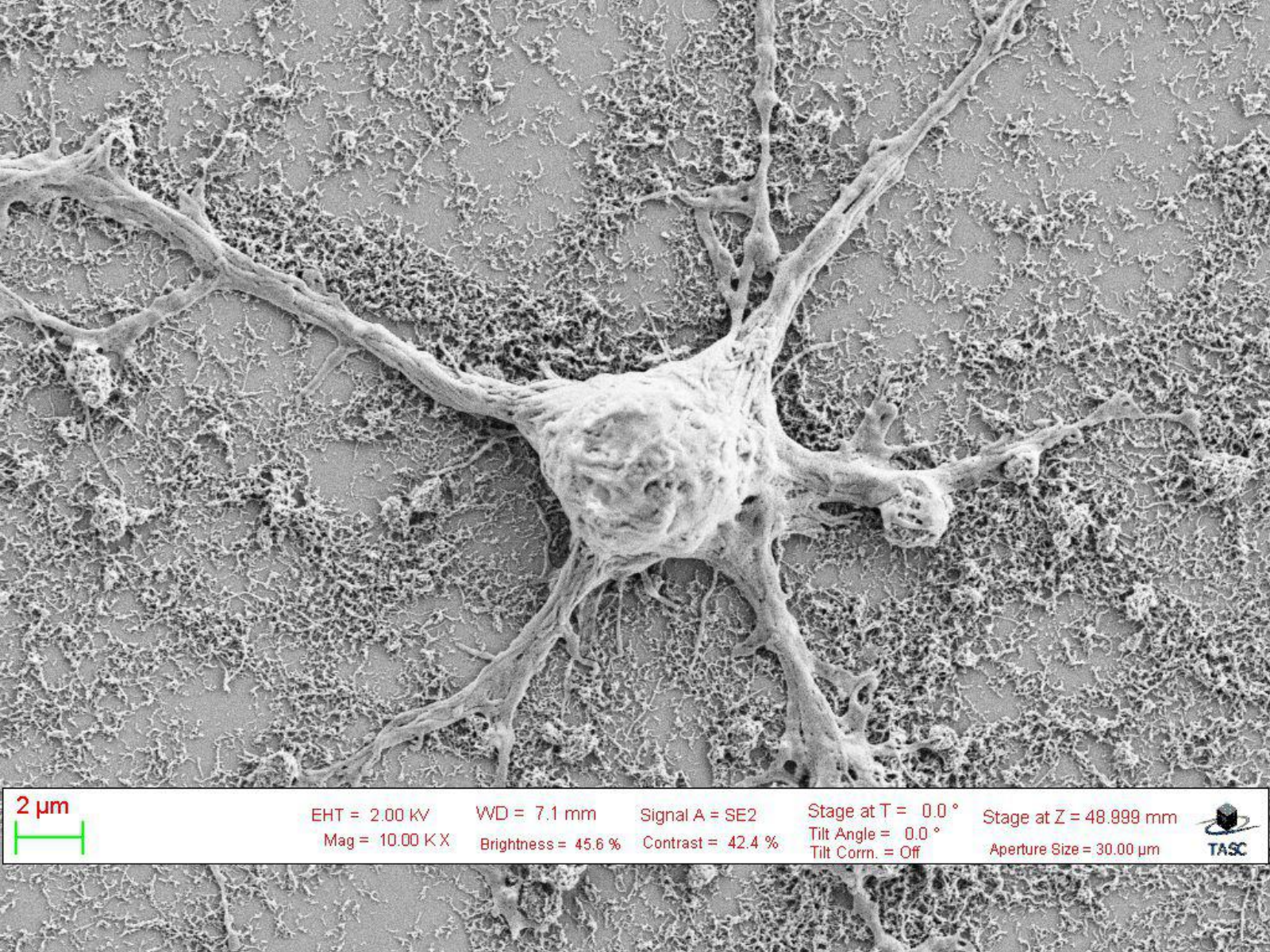} & The nanomaterials in the image have a dendritic, branching structure with a central node and multiple filament-like extensions. & The nanomaterials in the image possess a dendritic, branching structure with a central node and several filament-like extensions. & 0.824 & 0.895 & 0.944 \\ \hline
        \includegraphics[width=2cm,height=1.5cm,keepaspectratio]{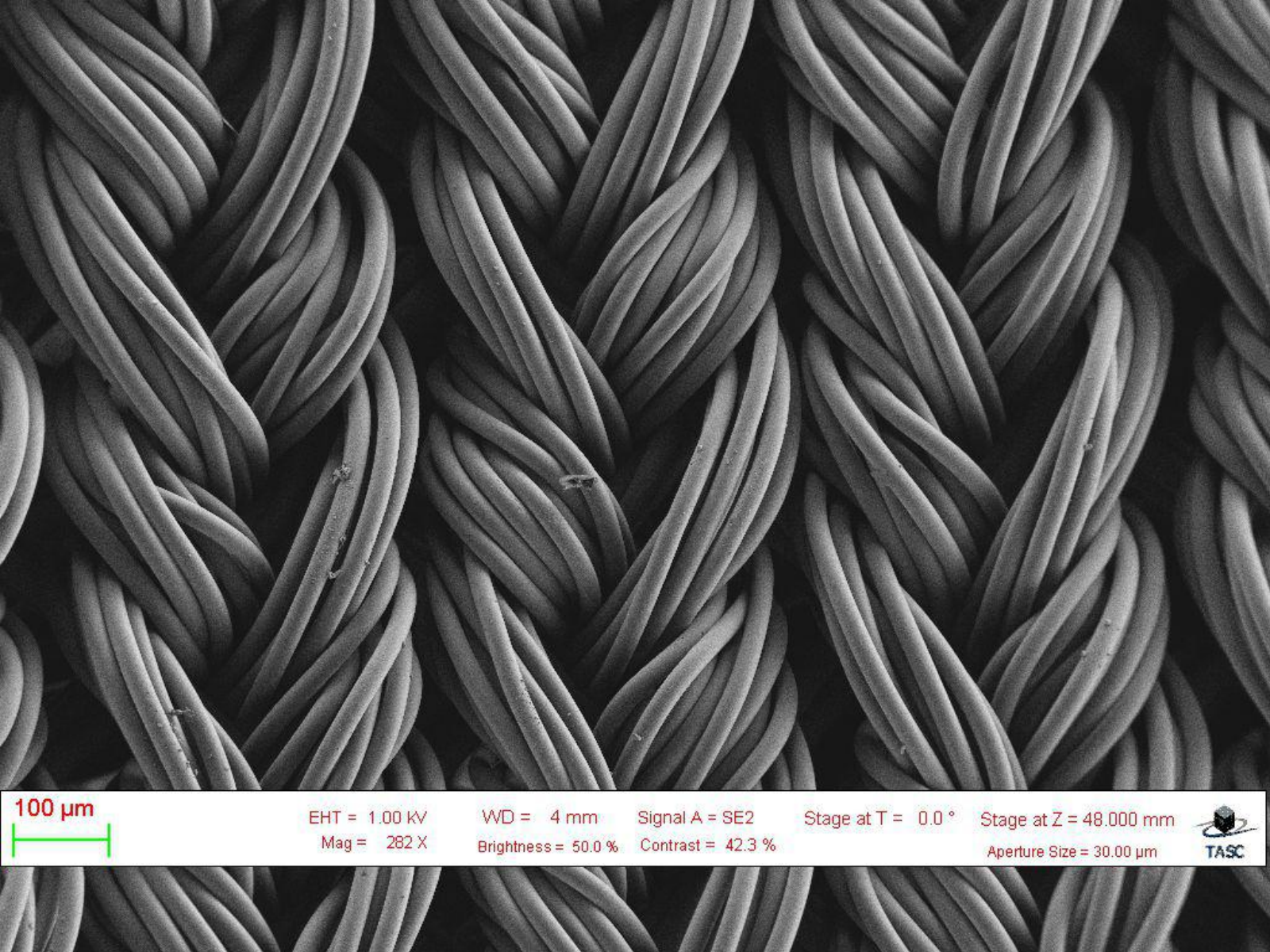} & The nanomaterials depicted resemble tightly woven, twisted cables or fibrous strands, densely packed and intertwined. & The nanomaterials depicted appear as tightly woven, twisted cables or fibrous strands, densely packed and interlaced. & 0.772 & 0.839 & 0.859 \\ \hline
        \includegraphics[width=2cm,height=1.5cm,keepaspectratio]{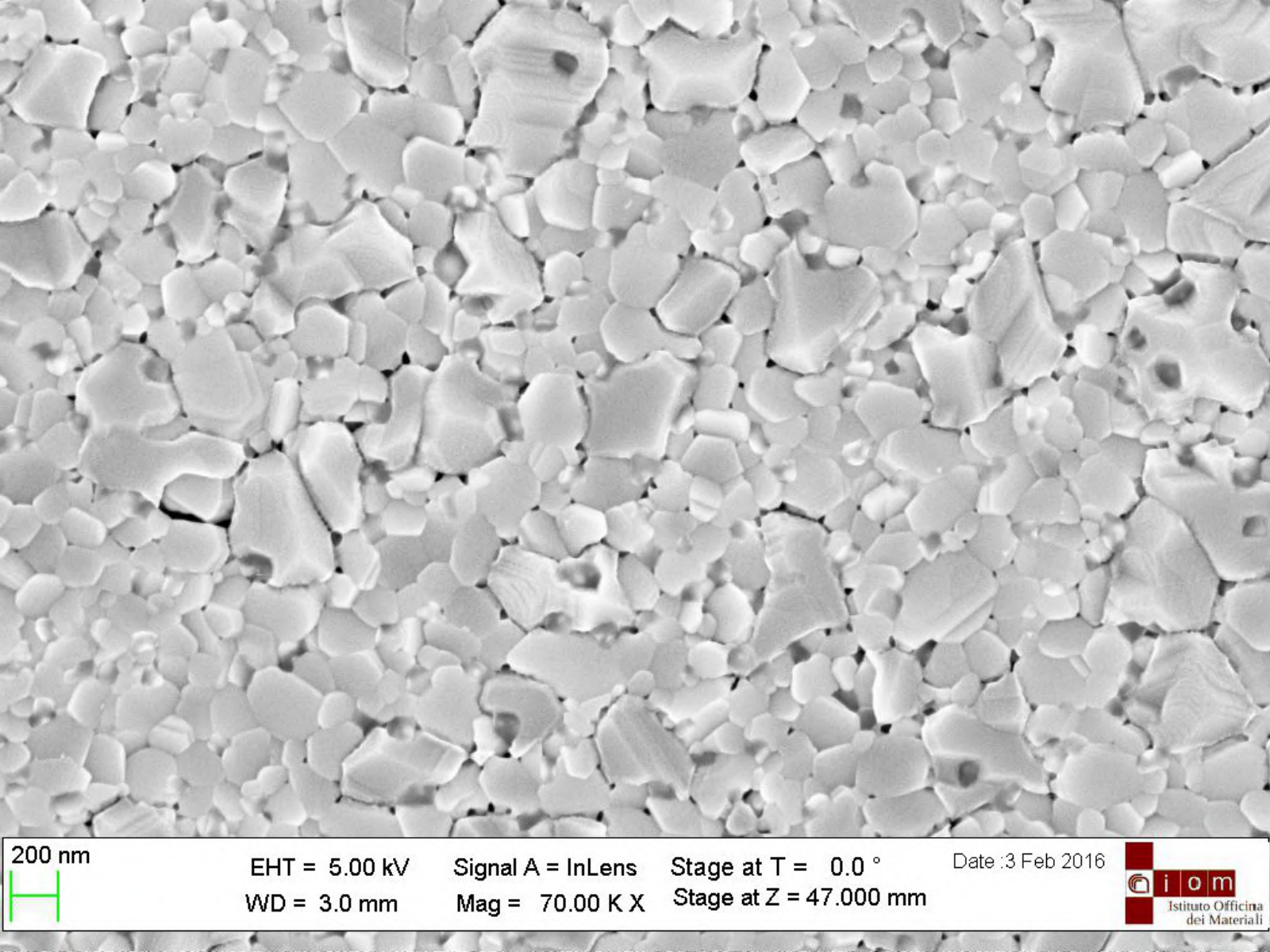} & The nanomaterials have a polygonal, plate-like morphology with irregular edges, giving them a shattered glass or cracked ice appearance. & The nanomaterials have polygonal, plate-like morphology with irregular edges, giving them a shattered glass or cracked ice appearance. & 0.918 & 0.974 & 0.952 \\ \hline
        \includegraphics[width=2cm,height=1.5cm,keepaspectratio]{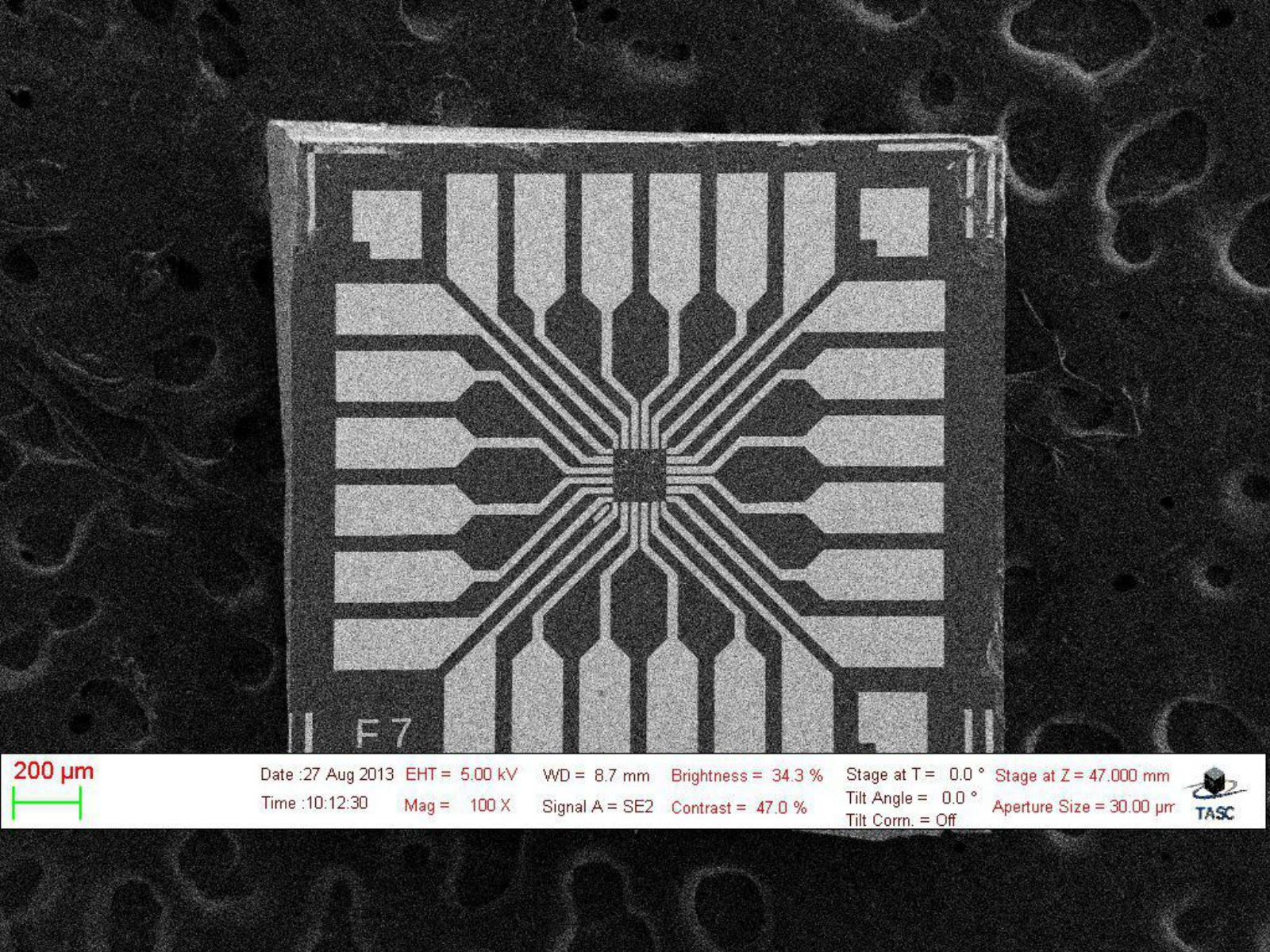} & The image depicts a square microfabricated device with uniform linear patterns on a granular semiconductor or nanoparticle substrate. & The image shows a square microfabricated device with uniform linear patterns on a granular semiconductor or nanoparticle substrate. & 0.913 & 0.944 & 0.999 \\ \hline
        \includegraphics[width=2cm,height=1.5cm,keepaspectratio]{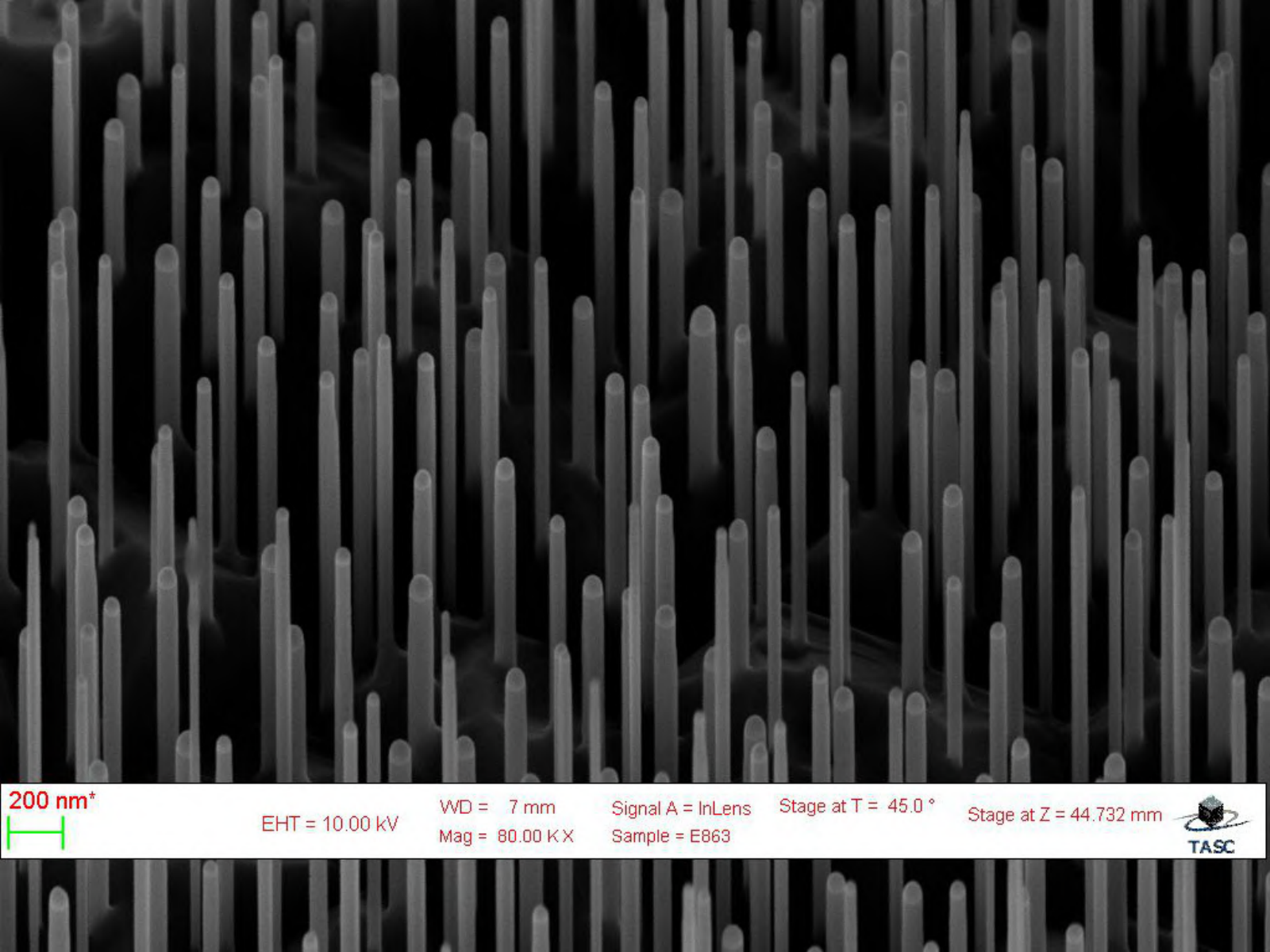} & The nanomaterials in the image exhibit a needle- or rod-like morphology, standing vertically and densely packed, similar to a bed of nails. & The nanomaterials in the image display a needle- or rod-like morphology, standing vertically and densely packed, akin to a bed of nails. & 0.858 & 0.913 & 0.954 \\ \hline
        \includegraphics[width=2cm,height=1.5cm,keepaspectratio]{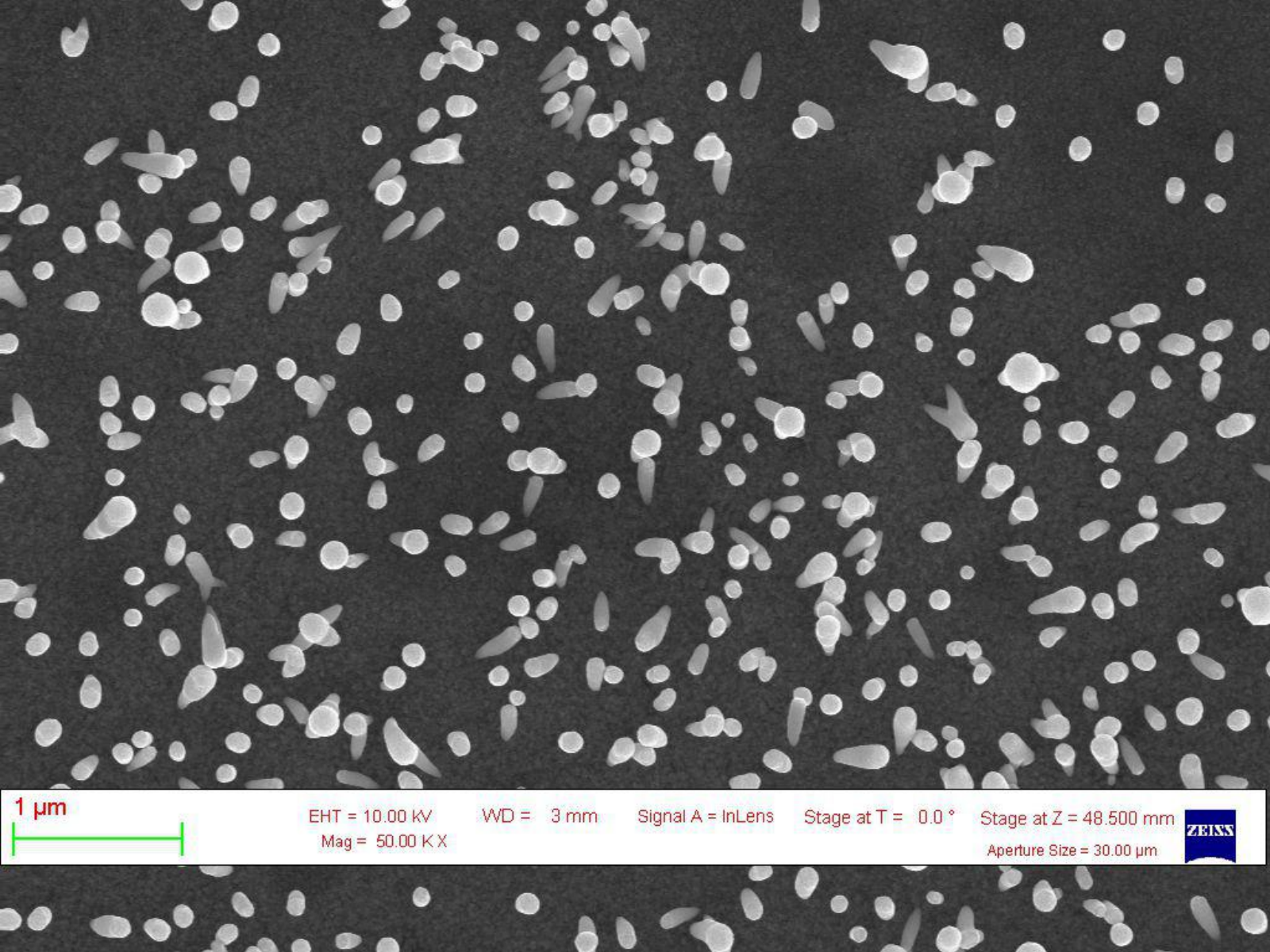} & The nanomaterials shown are elliptical or rod-shaped with smooth surfaces, scattered randomly across the surface. & The nanomaterials displayed are elliptical or rod-shaped with smooth surfaces, dispersed randomly across the surface. & 0.787 & 0.875 & 0.861 \\ \hline
        \includegraphics[width=2cm,height=1.5cm,keepaspectratio]{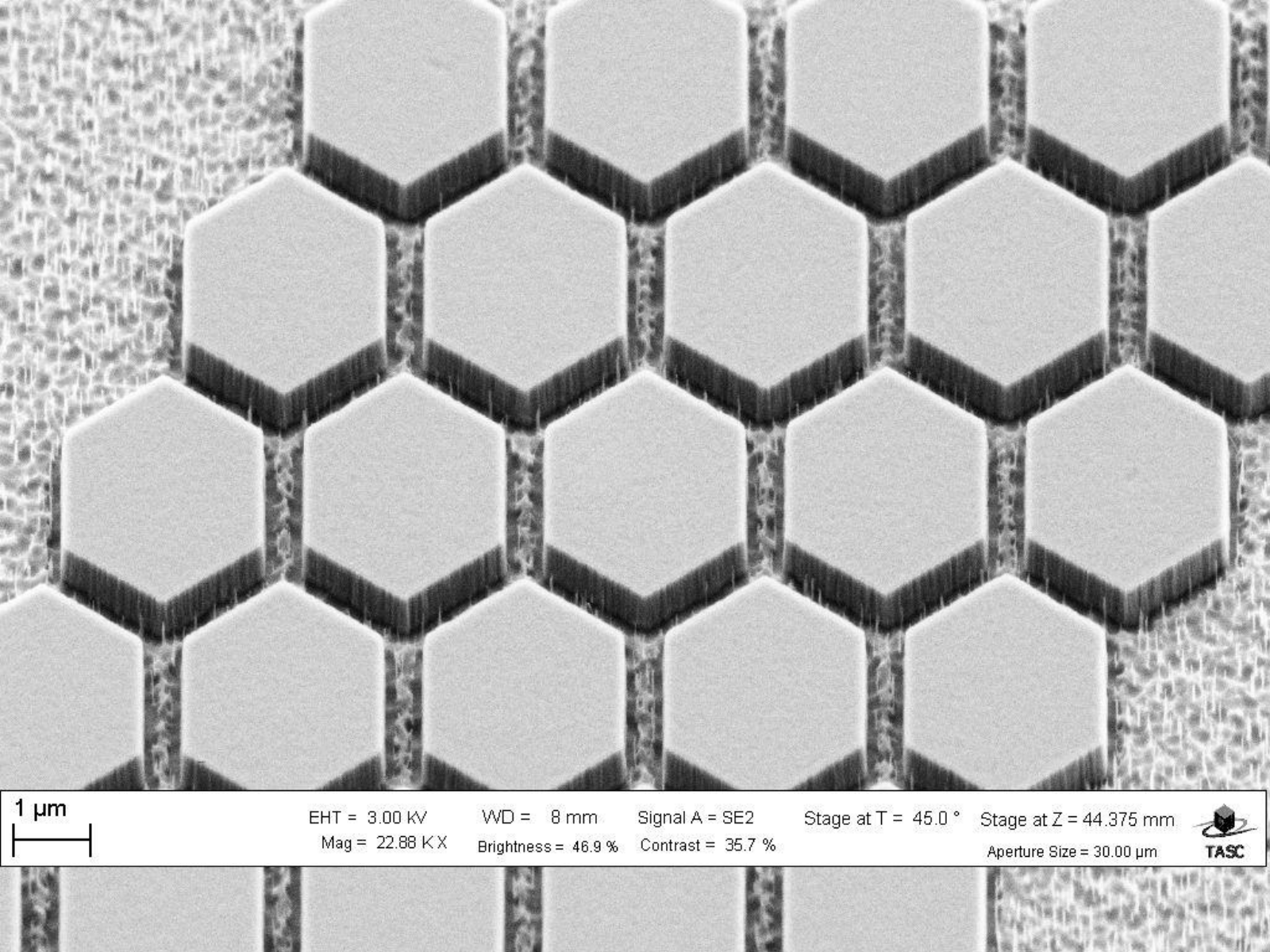} & The nanomaterials have a hexagonal, honeycomb-like structure, organized in a highly ordered, tessellated pattern. & The nanomaterials display a hexagonal, honeycomb-like structure, organized in a highly ordered, tessellated pattern. & 0.886 & 0.933 & 0.927 \\ \hline
        \includegraphics[width=2cm,height=1.5cm,keepaspectratio]{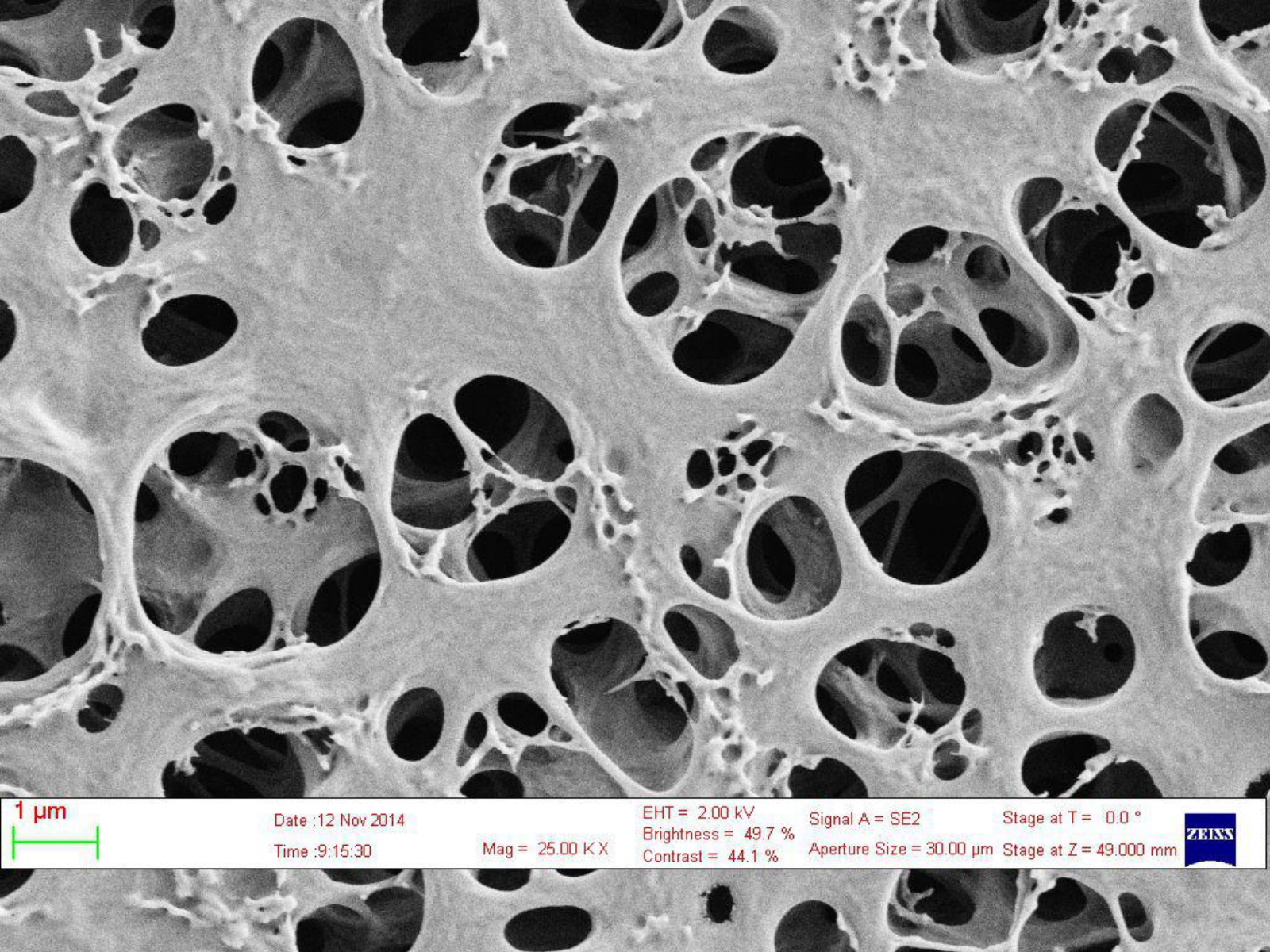} & The nanomaterials exhibit a foam-like structure with a network of interconnected pores of various sizes and irregular shapes, creating a porous, sponge-like morphology. & The nanomaterials display a foam-like structure with a network of interconnected pores of various sizes and irregular shapes, forming a porous, sponge-like morphology. & 0.820 & 0.920 & 0.913 \\ \hline
        \includegraphics[width=2cm,height=1.5cm,keepaspectratio]{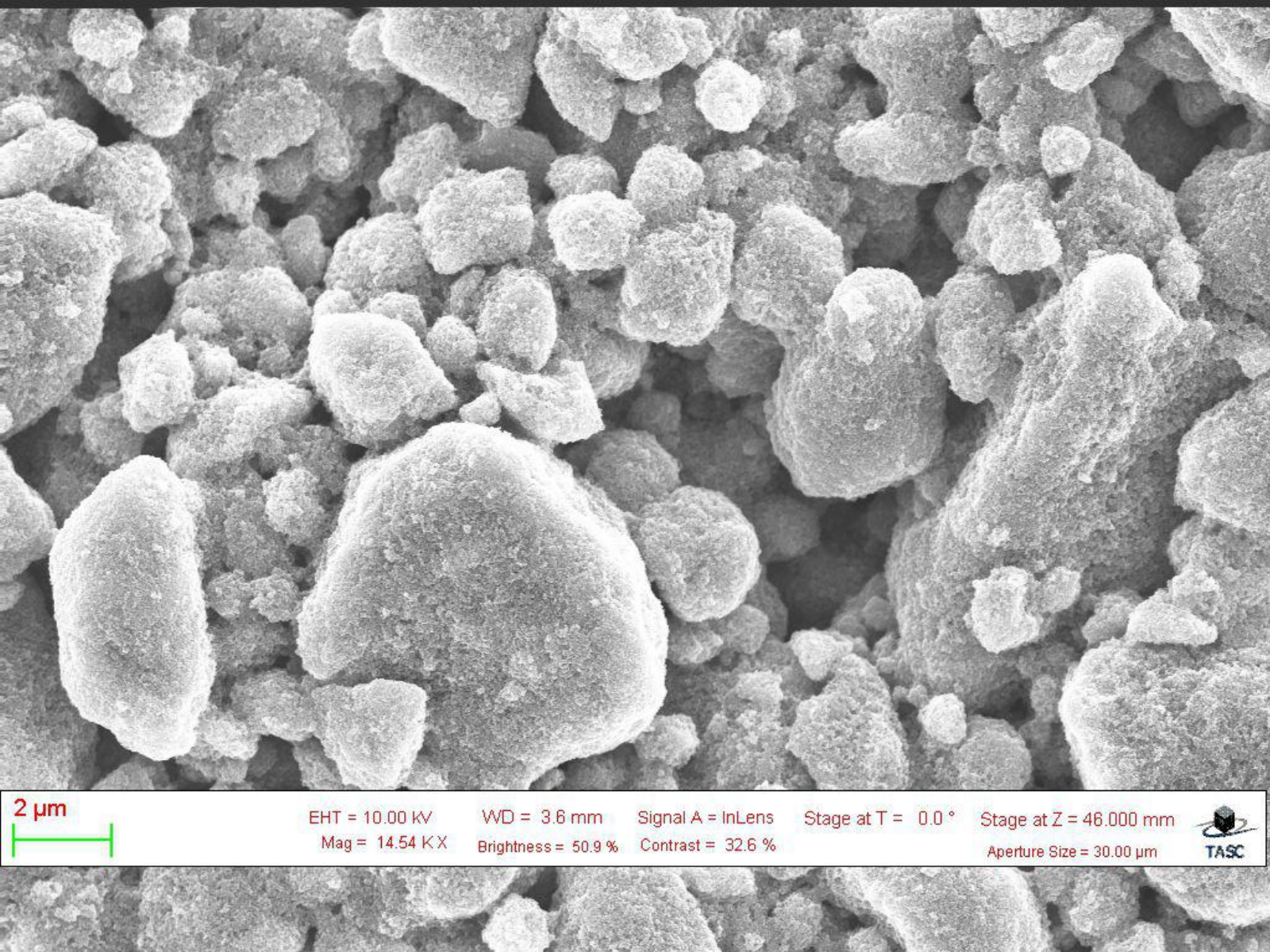} & The nanomaterials are irregularly shaped, resembling clumped aggregates with a rough, textured surface. & The nanomaterials appear irregularly shaped, resembling clumped aggregates with a rough, textured surface. & 0.877 & 0.920 & 0.920 \\ \hline
        \includegraphics[width=2cm,height=1.5cm,keepaspectratio]{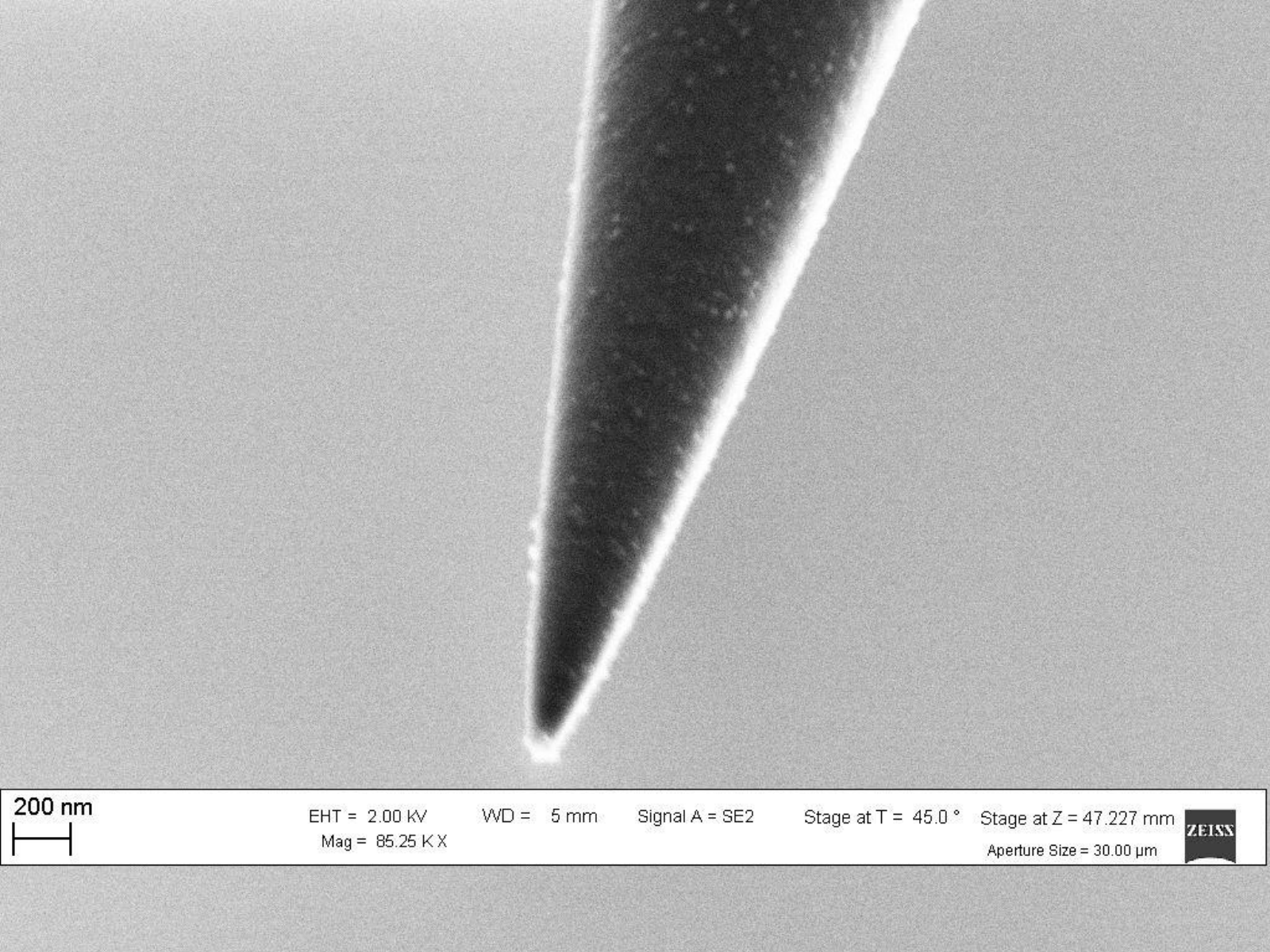} & The nanomaterial appears as a sharply pointed, conical structure with a smooth surface, tapering to a fine tip. & The nanomaterial is seen as a sharply pointed, conical structure with a smooth surface, tapering to a fine tip. & 0.863 & 0.920 & 0.938 \\ \hline
        \end{tabular}    
\end{table*}

\vspace{-5mm}
\bibliography{aaai24}

\end{document}